\documentclass{article}

\usepackage{multirow}
\usepackage[normalem]{ulem}
\useunder{\uline}{\ul}{}
\usepackage{enumitem} 
\usepackage{inconsolata}
\usepackage{wrapfig}
\usepackage{amstext}
\usepackage{graphicx}
\usepackage{floatrow}
\newfloatcommand{capbtabbox}{table}[][\FBwidth]
\usepackage{wrapfig}
\usepackage{caption}

\usepackage{array}
\usepackage{fvextra}
\usepackage[T1]{fontenc}
\usepackage[utf8]{inputenc}

    \usepackage[preprint]{neurips_2023}

\usepackage[utf8]{inputenc} 
\usepackage[T1]{fontenc}    
\usepackage{hyperref}       
\usepackage{url}            
\usepackage{booktabs}       
\usepackage{amsfonts}       
\usepackage{nicefrac}       
\usepackage{microtype}      
\usepackage{xcolor}         

\usepackage{hyperref}       
\usepackage{url}            
\usepackage{booktabs}       
\usepackage{amsfonts}       
\usepackage{nicefrac}       
\usepackage{microtype}      
\usepackage{lipsum}
\usepackage{fancyhdr}       
\usepackage{graphicx}       
\usepackage{tabularx}       
\graphicspath{{media/}}     
\title{Code Prompting: a Neural Symbolic Method for Complex Reasoning in Large Language Models}

\author{%
Yi Hu \quad Haotong Yang \quad Zhouchen Lin \quad Muhan Zhang\\
Institute for Artificial Intelligence, Peking University\\
{\ttfamily huyi2002@stu.pku.edu.cn}\\
{\ttfamily \{haotongyang, zlin, muhan\}@pku.edu.cn}\\
}

\begin{document}

\maketitle

\renewcommand{\thefootnote}{\fnsymbol{footnote}} 


\begin{abstract}
    Large language models (LLMs) have scaled up to unlock a wide range of complex reasoning tasks with the aid of various prompting methods. However, current prompting methods generate natural language intermediate steps to help reasoning, which can cause imperfect task reduction and confusion. To mitigate such limitations, we explore code prompting, a neural symbolic prompting method with both zero-shot and few-shot versions which triggers code as intermediate steps. We conduct experiments on 7 widely-used benchmarks involving symbolic reasoning and arithmetic reasoning. Code prompting generally outperforms chain-of-thought (CoT) prompting. To further understand the performance and limitations of code prompting, we perform extensive ablation studies and error analyses, and identify several exclusive advantages of using symbolic promptings compared to natural language. We also consider the ensemble of code prompting and CoT prompting to combine the strengths of both. Finally, we show through experiments how code annotations and their locations affect code prompting.
\end{abstract}

\begin{figure}[t]
    \centering
    \includegraphics[width=\textwidth]{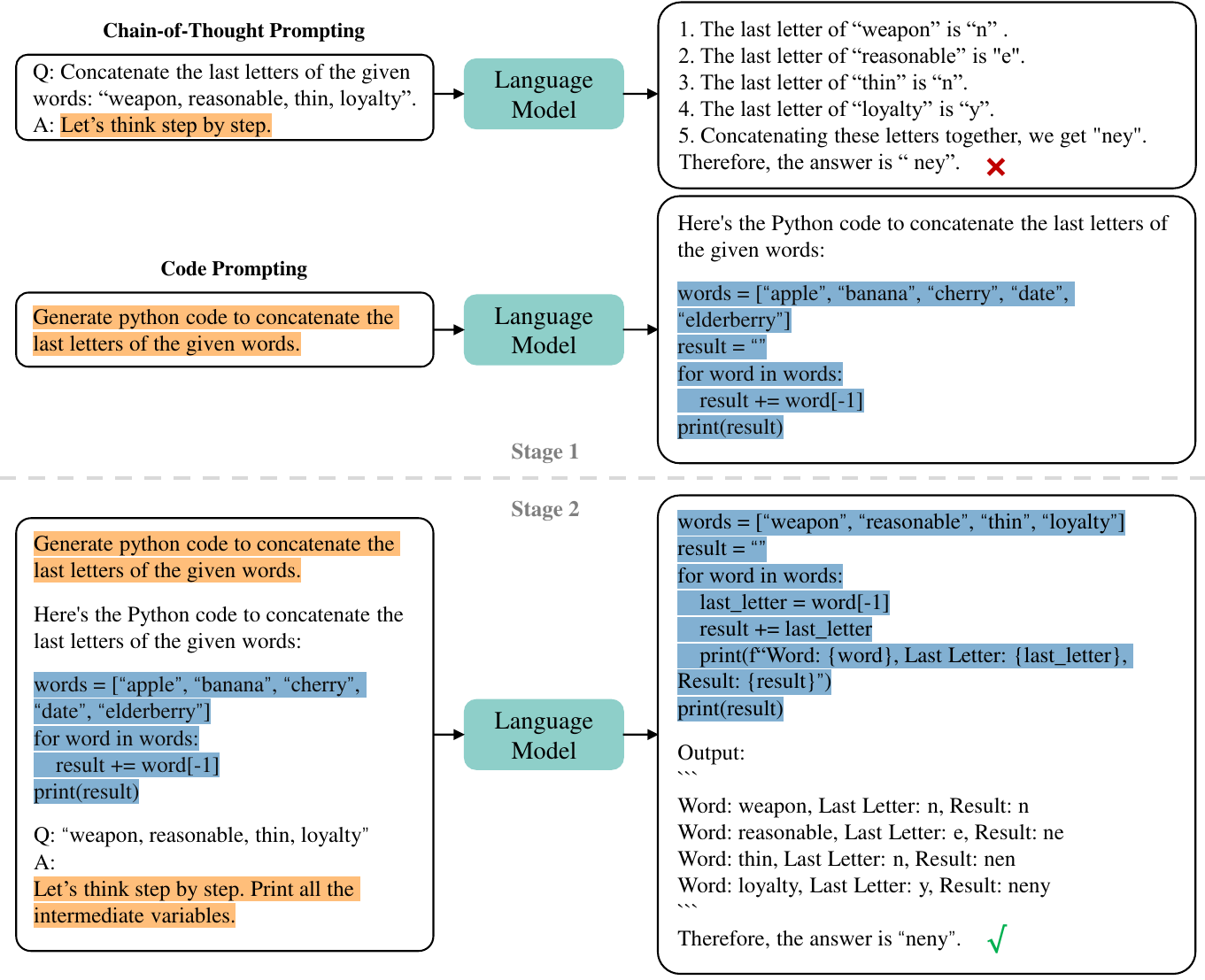}
    \caption{The pipelines of zero-shot CoT prompting and zero-shot code prompting.}
    \vspace{-5pt}
    \label{fig:pipeline}
\end{figure}



\section{Introduction}

Recent years have seen huge revolutions in the field of Natural Language Processing (NLP) regarding the shockingly fast development of large language models (LLMs)~\citep{gpt-3,gpt-4,llama,opt,lamda,palm,t5}. The scaling law~\citep{scaling_law} suggests a promising way to enhance a language model by scaling up the model size. 
According to~\citet{chatgpt_general,gpt-4}, LLMs have shown certain levels of the ability to solve complex reasoning tasks. Besides the size of LLMs, ``prompt engineering'' is crucial to the reasoning ability. Prompting methods like chain-of-thought prompting~\citep{zero-shot_cot,few-shot_cot} and least-to-most prompting~\citep{ltm} lead LLMs to generate rationales before the final answer and unlock a wide range of complex reasoning tasks. However, the model performance is highly dependent on the prompt design. How to craft simple, effective and general-purpose prompts still remains a hotly-debated open question.

A large number of works have proposed different prompting methods to facilitate LLM reasoning~\citep{few-shot_cot,auto_cot,zero-shot_cot,ltm, complexity_based_prompting,decomposed_prompting,self_ask}. These prompting methods induce LLMs to generate natural language intermediate steps before the final answer and enhance the reasoning ability of LLMs to a great extent. However, there still remain limitations.
First, although these methods lead LLMs to solve the problems step by step, they fail to plan the whole process of solution ahead, which is likely to result in unreasonable task reduction. In other words, the ``step'' can be too big for the LLM to stride over. For example, in Figure~\ref{fig:pipeline}, we show a case where the LLM is supposed to concatenate the last letters of the given words. Chain-of-thought prompting leads the LLM to first extract the last letter of each word and then concatenate all the letters together \textbf{in one step}. However, the step may be too difficult for the LLM with the number of letters increasing, resulting in wrong answers. Second, ambiguity naturally lies in natural language. Rationales of natural language may also cause confusion, thus misguiding the reasoning process. 
 
To mitigate the limitations of the above methods, we consider combining neural methods with symbolic ones, as symbolic methods have structured, unambiguous representations and computation flows, which may make the reasoning process more accurate and alleviate the effect of ambiguity. Specifically, we induce neural methods (LLMs) to \textbf{first generate symbolic solvers} (such as computation graphs) and let the symbolic solvers to \textbf{output the final answers}. Prior work of neural-symbolic methods constructs intermediate steps by formal languages instead of natural language~\citep{ns:amini2019mathqa,ns:chen2020compositional,ns:Chen2020Neural,ns:chiang2019semanticallyaligned,ns:roy-etal-2015-reasoning,declarative}. However, they are often limited to a specific field 
and have trouble generalizing to broader reasoning tasks. 

To overcome the difficulties, recent works~\citep{pal,pot,binder,self-debug} have tried to facilitate complex reasoning in LLMs with programs/code. As a general-purpose tool, code is machine-executable, LLM-producible, and can define flexible computation graphs. We follow the works and further explore code prompting, a neural-symbolic way to enhance the reasoning ability of LLMs. As the corpus on which LLMs are pre-trained expands to contain code, LLMs are naturally endowed with the ability to generate code following certain instructions~\citep{codex}. We expect to leverage the code generation ability of LLMs to promote reasoning. Our code prompting is a two-stage method with the pipeline shown in Figure~\ref{fig:pipeline}. To solve a specific question, code prompting first guides LLMs to generate a piece of code. To fully exploit the reasoning ability of pure LLMs, we explore two ways of leveraging code prompting. One is to instruct LLMs to follow the generated code to complete the reasoning process and finally give the answer. The other is to call an external compiler to execute the code. 

We find out in experiments that code 
works as a mind map for the LLM and reduces the task into sub-tasks represented by separate operations in the code. As the separate operations in code are often easy for LLM, the task reduction greatly facilitates LLMs to solve the complex reasoning problems step by step by themselves. Besides, code is a formal language without ambiguity, which eliminates confusion caused by ambiguity in natural language and makes the reasoning process more robust. Regarding the limitations of prior neural-symbolic methods, code is a formal language with the potential to adapt to a wide range of tasks.

We evaluate code prompting on 7 datasets involving symbolic and arithmetic reasoning. We show that for symbolic reasoning, zero-shot code prompting outperforms zero-shot chain-of-thought prompting by at least 26.2\% (71.0\% $\rightarrow$ 97.2\%) in the last letter concatenation task and by 11.7\% (75.3\% $\rightarrow$ 87.0\%) in average in the task of coin flip even without the help of external tools like an Python interpreter. For arithmetic reasoning, zero-shot code prompting generally outperforms zero-shot chain-of-thought prompting, and few-shot code prompting is highly competitive to the baselines. 

We further dig into the failure cases of code prompting. Through error analysis, we identify several limitations of the method in addition to some key aspects of its success. Moreover, we find out that code prompting and chain-of-thought prompting can lead the LLM to think from different angles, suggesting a combination of both methods, which achieves 87.95\% accuracy on GSM8K.

\section{Related Work}
\textbf{Prompt engineering.}
Recently, various prompting methods have emerged to boost LLM reasoning~\citep{few-shot_cot,auto_cot,zero-shot_cot,ltm, complexity_based_prompting,decomposed_prompting,self_ask}. They have shown effectiveness in a wide range of reasoning tasks. Classified based on whether specific exemplars are provided, prompting methods can be divided into zero-shot prompting and few-shot prompting. For example, the widely-used technique chain-of-thought prompting has a zero-shot version~\citep{zero-shot_cot} and a few-shot version~\citep{few-shot_cot}. Few-shot prompting calls for manually constructed task-specific demonstrations, so it is more costly. \citet{auto_cot} introduced a way to automatically construct few-shot demonstrations based on zero-shot prompting to save the manual cost. In this paper, we introduce both zero-shot and few-shot code prompting and show the competitive results of both versions regarding symbolic and arithmetic reasoning. Existing prompting methods can also be divided into single-step prompting and multi-step prompting methods. CoT prompting~\citep{zero-shot_cot,few-shot_cot} is a classic single-step prompting method. A recent work~\citep{ltm} has introduced least-to-most prompting, which is a multi-step prompting method that first divides a question into several sub-questions and then guides the LLM to solve each question sequentially. Least-to-most prompting is limited to a few-shot version and requires careful hand-crafting of demonstrations which vary significantly from task to task. Thus, it is less general than CoT and our code prompting. 

\textbf{Program-aided reasoning.} Prior works have shown that programs generated by LLMs may facilitate reasoning~\citep{pal,pot,palm}. A recent work demonstrates a similar idea of using code to elicit LLM reasoning, which is called Program-aided Language Models (PAL)~\citep{pal}. We follow the work and dig deeper into code prompting: 1) PAL is a few-shot prompting method, while we investigate both zero-shot and few-shot code prompting. We show that zero-shot code prompting matches or even exceeds PAL in arithmetic reasoning tasks. Besides, we improve the few-shot performance too. 2) We offer another option on the second stage of final answer generation, namely directly asking LLM to generate the final answer according to the code, instead of calling a Python interpreter as in~\citet{pal}. Our experiments on using LLM in the second stage further prove that code prompting indeed assists reasoning in LLM even without executing the code and that symbolic and arithmetic reasoning abilities are self-contained for LLMs. 3) We discuss some key limitations and insights of code prompting through error analysis. These points help us understand the power of code prompting better and motivate us to combine CoT and code prompting. 

\textbf{LLMs with external tools.} 
Code prompting has an option to adopt a Python interpreter as an external tool to assist the LLM to complete the tasks. The concept of augmenting LLMs with external tools has drawn much attention~\citep{decomposed_prompting,binder,self_ask}. A similar work~\citep{binder} also uses programs to assist LLM reasoning. However, they focus on generating SQL or SQL-like programs to deal with questions of reasoning with tables. Besides, a recent work~\citep{self-debug} introduces a method of teaching LLMs to debug their generated program with the help of feedback from a code interpreter. In our work, we equip code prompting with a similar technique. However, we consider reasoning tasks while \citet{self-debug} focuses on tasks of text-to-code or code-to-code generation.

\section{Code Prompting}
\label{sec:methods}

Code prompting guides LLMs to solve a complex problem in two stages:
\begin{enumerate}[leftmargin=*]
    \item \textbf{Code Generation.}
    In the first stage, the prompt asks the LLM to generate Python code to solve the problem. We test on both zero-shot and few-shot prompting. The zero-shot prompt is simply ``Generate python code to \(\left\{ \right.\)task description\(\left. \right\} \).''. The few-shot prompt consists of exemplars of questions followed by corresponding code.
    \item \textbf{Based-on-code Solution.}
    The second stage is for solving the problem on the basis of the code generated in the first stage. We compare two different methods. The first method is to still ask the LLM to generate rationales and the solution to the problem according to the code. The second method is to directly use an external Python interpreter to execute the code. 
\end{enumerate}


In Figure~\ref{fig:pipeline}, we show the pipeline of zero-shot code prompting in the task of last letter concatenation introduced by~\citet{few-shot_cot}, which is a commonly used benchmark for symbolic reasoning~\citep{few-shot_cot,zero-shot_cot,ltm,auto_cot}. In the first stage, we use the zero-shot prompt to guide the LLM to output a piece of code for the task. To be more specific, we prompt the LLM with the instruction ``Generate python code to concatenate the last letters of the given words.''. In the second stage, the LLM is given the code it generates in the first stage and the specific question. Additionally, we add the sentence ``Let's think step by step. Print all the intermediate variables.'' to the prompt to further elicit the reasoning ability of the LLM. 

As mentioned before, the first stage (code generation) has the option of few-shot prompting instead of zero-shot prompting, and the second stage (solution) has the option of using an external code interpreter instead of prompting the LLM. They are not shown in the figure. 

\begin{figure}[t]
    \centering
    \includegraphics[width=\textwidth]{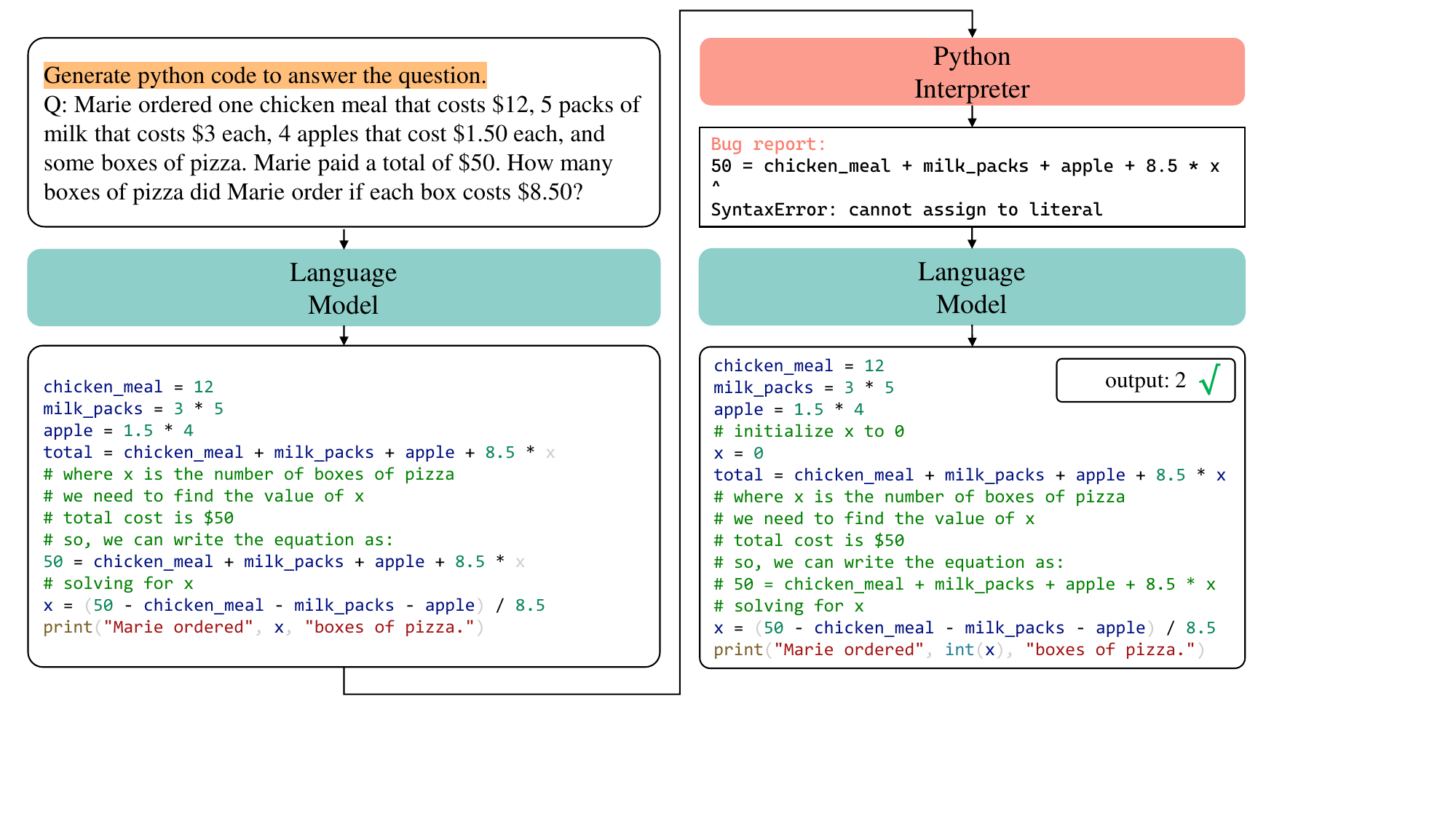}
    \caption{The pipeline of ``self-debugging''.}
    \label{fig:self-debug}
\vspace{-5pt}
\end{figure}

\paragraph{Self-debugging}
\label{sec:self-debug}
If a python interpreter is used in the second stage, we can equip code prompting with a ``self-debugging'' module. That is, when the interpreter returns errors (not often, the code is runnable in most cases), we simply append the generated code (with bugs) and the bug report to the original model input to prompt the LLM again. The pipeline of the ``self-debugging'' module is shown in Figure~\ref{fig:self-debug}. We show in experiments that state-of-the-art LLMs are capable of learning from the bug report and fixing the code.

In the following experiments, we show the results of code-prompting for symbolic reasoning and arithmetic reasoning. We conduct all the experiments on the model gpt-3.5-turbo (ChatGPT) and set the temperature to 0 unless otherwise specified.

\section{Symbolic Reasoning}
\label{sec:symbolic}

\subsection{Tasks}
We consider two symbolic reasoning tasks introduced by~\citet{few-shot_cot} and widely used in~\citet{few-shot_cot,zero-shot_cot,ltm,auto_cot}.

\textbf{Last letter concatenation.} The task asks LLMs to concatenate the last letters of given words. We follow \citet{ltm} to construct word lists by randomly selecting words from five thousand words of the Wikipedia frequency list. We construct word lists of lengths 4, 8 and 12. For each length, we test the prompting methods on 500 word lists, which form a test dataset of 1,500 samples. 

\textbf{Coin flip.} The task requires LLMs to answer whether a coin is still heads up after several people flipped or did not flip it. The number of people varies from 3 to 5. For each certain number of people, we construct 500 questions, which form a test dataset of 1,500 samples. 

\subsection{Methods}\label{sec:symbolic_methods}
We consider zero-shot standard prompting and zero-shot chain-of-thought prompting as baselines. For our code prompting methods, we only consider zero-shot code prompting because the code for each question is basically the same so that the exemplars in few-shot code prompting may leak the answer. We also show the results of few-shot chain-of-thought prompting for reference.

\textbf{Zero-shot Standard Prompting.} Zero-shot standard prompting has the form of ``Q: \(\left\{ \right.\)\textit{question}\(\left. \right\}\)$\backslash$nA:''. This prompting method reflects the symbolic reasoning ability of the LLM without any prompt engineering.

\textbf{Zero-shot Chain-of-Thought Prompting.} Zero-shot chain of thought prompting (zero-shot CoT prompting) is introduced by~\citet{zero-shot_cot}. As shown in Figure~\ref{fig:pipeline}, ``Let's think step by step.'' is appended to the question to elicit the LLM to generate chains of thought leading to the final answer.

\textbf{Few-shot Chain-of-Thought Prompting.} Few-shot chain of thought prompting (few-shot CoT prompting) is introduced by~\citet{few-shot_cot}. We provide 2 exemplars for the task of last letter concatenation and 8 exemplars for the task of coin flip. See Appendix~\ref{appendix:exp_symbolic} for the demonstrations.

\textbf{Zero-shot Code Prompting + LLM Self-contained.} As is shown in Figure~\ref{fig:pipeline}, our proposed method prompts the LLM to first generate task-specific code and then follow the code to generate the final answer. For zero-shot code prompting + LLM self-contained, we use the LLM itself to perform reasoning in the second stage. For the task of last letter concatenation, the prompt of the first stage is ``Generate python code to concatenate the last letters of the given words.''; for the task of coin flip, the first-stage prompt is ``A coin is heads up, there are some people, each one flipped or didn’t flip the coin. Generate python code to determine whether the coin is still heads up.'' The prompt we use in the second stage is ``Q: \(\left\{ \right.\)question\(\left. \right\} \)$\backslash$nA: Let's think step by step. Print all the intermediate variables.''

\textbf{Zero-shot Code Prompting + Python Interpreter.} 
The first stage of the method is the same as zero-shot code prompting + LLM self-contained. In the second stage, we employ a Python interpreter to execute the code to get the final answer directly in the second stage.

\subsection{Results}

As shown in the Table~\ref{table:symbolic}, even without external tools like Python interpreter, code prompting outperforms the baseline prompting methods (here we mainly consider zero-shot CoT prompting) by at least 26.2\% in the task of last letter concatenation and by 11.67\% on average in the task of coin flip. Besides, the performance gain increases with the complexity of the questions, implying that code prompting has better generalization ability in terms of symbolic reasoning. Specifically, the performance gain grows from 26.2\% to 72.8\% in the task of last letter concatenation when the length of word lists increases from 4 to 12, and in the task of coin flip, the performance gain rises from 0.0\% to 18.0\% with flip times increasing from 3 to 5. Compared to few-shot CoT prompting, zero-shot code prompting + Python interpreter generally performs better. Besides, it is noteworthy that zero-shot code prompting + LLM self-contained achieves better performance in the task of last letter concatenation without the help of Python interpreter. Even though few-shot CoT prompting outperforms code prompting + LLM self-contained in the task of coin flip, it should be noticed that few-shot methods are more costly due to the need for task-specific manual crafting of demonstrations.

\begin{table}[t]
\centering
\begin{tabular}{cccccccc}
\toprule[1pt]
    & \multicolumn{3}{c}{last letter concatenation} &  & \multicolumn{3}{c}{coin flip}  \\ 
 & l=4 & l=8 & l=12 &  & \#flips=3 & \#flips=4 & \#flips=5 \\\midrule
zero-shot standard & 7.4 & 2.0 & 2.6 &  & 22.6 & 17.2 & 16.8     \\
zero-shot CoT & 71.0 & 27.8 & 2.8 &  & 86.2 & 71.8 & 67.8     \\
few-shot CoT &  94.6 &  69.6 &  39.8 &  & \textbf{99.8} & \textbf{99.8} & {\ul 99.0}     \\
\midrule
zero-shot code + LLM self-contained & {\ul 97.2} & {\ul 85.4} & {\ul 75.6} &  &  86.2 &  88.8 &  85.8   \\
zero-shot code + Python interpreter & \textbf{99.4} & \textbf{99.8} & \textbf{99.8} && \textbf{99.8} & \textbf{99.8} & \textbf{99.4} \\

\bottomrule[1pt]
\end{tabular}
\setlength{\abovecaptionskip}{0.3cm}
\caption{The accuracy (\%) of zero-shot standard prompting, zero-shot CoT prompting, few-shot CoT prompting and zero-shot code prompting on last letter concatenation and coin flip. 
}
\label{table:symbolic}
\end{table}

We summarize the possible reasons for the outstanding performance of code prompting as follows:
\begin{enumerate}[leftmargin=*]
    \item \textbf{Abstraction and Simplification.} Code prompting urges the LLM to generate code from the question, which can be seen as a step of abstraction and simplification. For example, in both tasks, code prompting endows the LLM the ability to extract the ``loop'' nature of the question and leverages the nature explicitly in the code using ``for'' or ``while'' syntax in Python language. The LLM might benefit from the explicit identification of loop and switch to its recursive mode.  
    \vspace{-3pt}
    \item \textbf{Task Reduction.} Code prompting transforms a question into a program. Each line of the code can be seen as a sub-task. Sub-tasks that can be represented as a line of code are usually easy for the LLM to complete, and thus the question transformation actually generates friendly instructions for LLM reasoning. For example, as shown in Figure~\ref{fig:pipeline}, in the task of last letter concatenation, CoT prompting guides the LLM to find the last letter of each word first and then concatenate them together. However, concatenating all letters in one step may be too difficult for the LLM when the number of letters is large. Meanwhile, code prompting leads the LLM to store the current result in a \texttt{string} and reduces the task to steps of extracting the last letter of a word and concatenating it to the \texttt{string} sequentially. Code prompting triggers a more reasonable task reduction, as the difficulty of each step is lower and does not grow with the number of words.
    \vspace{-3pt}
    \item \textbf{Disambiguation.}\label{disambiguation} The code generated by the LLM can be seen as a formalization of the question that eliminates the ambiguity in the original natural language questions. In contrast, other prompting methods like CoT may lead to confusion in the reasoning process. See Appendix~\ref{appendix:error_symbolic} for examples.
    \vspace{-3pt}
    \item \textbf{Explicit Template.} Through case-by-case observation, we discover that code generated in the first stage works as an explicit template, a mindmap for the LLM to conduct the reasoning process. LLM follows the code to complete the task. In comparison, CoT prompting fails to provide such an explicit template before reasoning, which proves in the experiments to be crucial.
\end{enumerate}

\section{Arithmetic Reasoning}
\label{sec:arithmetic}

\subsection{Tasks}
    We consider five commonly used datasets for the task of arithmetic reasoning.
    (1) SinglEq~\citep{singleEq}, (2) AddSub~\citep{addsub}, (3) MultiArith~\citep{multiarith}, (4) SVAMP~\citep{SVAMP}, (5) GSM8K~\citep{gsm8k}. 
    SingleEq, AddSub and MultiArith are from the MAWPS (MAth Word ProblemS) benchmark introduced by~\citet{mawps}. 
    Among the datasets, SingleEq and AddSub only take single-step calculation, while MultiArith, GSM8K and SVAMP contain harder math problems that require multi-step reasoning.
    See Appendix~\ref{appendix:data_arithmetic} for more details of each dataset.
    
\subsection{Methods}
\textbf{Zero-shot Chain-of-Thought Prompting.} This method is the same as in~\S\ref{sec:symbolic_methods}.

\textbf{Zero-shot Code Prompting.}\label{zero-shot code}
In the first stage, the prompt is ``Generate python code to answer the question.$\backslash$nQ:\(\left\{ \right.\)\textit{question}\(\left. \right\} \)''. In the second stage, we use an external Python interpreter. 
We additionally design several modules to augment code prompting, including \textbf{self-debugging}, \textbf{irrelevant information} (irr) and \textbf{equation instruction} (equ). The self-debugging is described in~\S\ref{sec:self-debug}. For irrelevant information, we simply add an instruction ``There may be irrelevant information in the question. If you find it, ignore it.'' to the zero-shot prompt. For equation instruction, we provide an instruction on solving equations in Python using the package \texttt{sympy}. See Appendix~\ref{appendix:exp_arithmetic} for more details.

\textbf{Few-shot Chain-of-Thought Prompting.} 
We consider few-shot CoT prompting introduced by~\citet{few-shot_cot} as a baseline. The pipeline is shown in Appendix~\ref{appendix:exp_arithmetic}. We use 3 of the exemplars introduced by~\citet{few-shot_cot} for all the math world problems. See Appendix~\ref{appendix:exp_arithmetic} for the exemplars.

\textbf{Program-aided Language Models.} 
We also consider the recent work Program-aided Language Models (PAL) introduced by~\citet{pal} as a baseline. PAL is a few-shot prompting method that uses program as prompts. We run the method on gpt-3.5-turbo using the same 3 exemplars.

\textbf{Few-shot Code Prompting.} 
In the first stage, we use the same 3 questions as few-shot CoT prompting to guide the LLM to generate code. We add \textbf{annotations} to raise the readability and further strengthen the method. In the second stage, we use an external Python interpreter. We also add augmenting modules, which are basically the same as in zero-shot code prompting. See Appendix~\ref{appendix:exp_arithmetic} for more details. We also experiment on using LLMs to execute the code in the second stage, the results are shown in Appendix~\ref{appendix:exp_arithmetic}.


\begin{table}[t]
\centering
\begin{tabular}{rccccc}
\toprule
\multicolumn{1}{l}{}      & SingleEq       & AddSub         & SVAMP          & MultiArith     & GSM8K          \\\midrule
\multicolumn{1}{l}{{\ul \textit{\textbf{zero-shot methods}}}} &    &    &    &    &    \\
zero-shot CoT             & 93.11          & 86.08          & 78.20          & \textbf{96.67} & \textbf{81.58} \\
zero-shot code (+self-debug)       & \textbf{97.64} & \textbf{91.65}(+irr)          & \textbf{79.40} & \textbf{96.67} & 78.92(+equ) 
\\\midrule
\multicolumn{1}{l}{{\ul \textit{\textbf{few-shot methods}}}}  &   &  &    & &    \\
few-shot CoT & 95.47 & 90.63  & 77.10 & \textbf{98.50} & \textbf{82.11} \\
PAL  & \textbf{97.64} & 89.11 & 79.50 & 97.00 & 79.45          \\
few-shot code (+self-debug)  & 97.44 & \textbf{91.39}(+irr) & \textbf{79.60} & 97.33 & 80.59  \\
\bottomrule
\end{tabular}
\setlength{\abovecaptionskip}{0.3cm}
\caption{The accuracy (\%) of zero-shot CoT prompting, zero-shot code prompting, few-shot CoT prompting, PAL~\citet{pal} and few-shot code prompting on arithmetic reasoning.}
\label{table:arithmetic}
\end{table}

\subsection{Results}
\label{subsec:results}
    Table~\ref{table:arithmetic} presents the results. Note that few-shot code prompting in the table adds annotations at the end of each line of the code. 
    Zero-shot code prompting outperforms the baselines by 4.53\%, 5.57\% and 1.20\% in SingleEq, AddSub and SVAMP respectively. Nevertheless, zero-shot code prompting is 1.52\% behind zero-shot CoT prompting in GSM8K. Few-shot code prompting generally outperforms PAL with the aid of annotations, except for a gap of 0.20\% in SingleEq. Besides, few-shot code prompting outperforms few-shot CoT prompting by 1.97\% in SingleEq, 0.76\% in AddSub and 2.50\% in SVAMP. However, code prompting is 1.17\% and 1.52\% behind few-shot CoT prompting in MultiArith and GSM8K respectively. We will discuss the errors in~\S\ref{error analysis}.

    It is noteworthy that \textit{zero-shot} code prompting generates competitive performance to \textit{few-shot} CoT prompting and PAL. Compared with few-shot CoT prompting, zero-shot code prompting performs better in SingleEq, AddSub and SVAMP by 2.17\%, 1.02\% and 2.30\%. Compared with PAL, zero-shot code prompting performs better in AddSub, and the performance gap in other datasets is no more than 0.53\%. This demonstrates that with proper augmentations, zero-shot code prompting is close to few-shot performance.

\subsection{Error analysis}
\label{error analysis}

\textbf{AddSub.}\label{error:addsub}
    In AddSub, we find that the code generated by the LLM tend to contain irrelevant variables, thus disturbing it. See Appendix~\ref{appendix:error_arithmetic} for examples. The observation suggests code generation by LLMs is easy to be distracted by irrelevant information. This motivates the irr augmentation.

\textbf{MultiArith.}
In MultiArith, the wrong answers from code prompting are mainly due to two reasons: 1) Code prompting is more sensitive to ambiguity in the question. See Appendix~\ref{appendix:error_arithmetic} for examples. (The feature of code prompting is discussed in detail in~\S\ref{discussion}). 2) The groundtruth for some questions is wrong, while the predictions are correct. Examples are listed in Appendix~\ref{appendix:error_arithmetic}.

\textbf{GSM8K.}\label{error:gsm8k} 
    For the dataset GSM8K, we observe that the LLM struggles to solve equations in code. 
    See error examples in Appendix~\ref{appendix:error_arithmetic} and Appendix~\ref{appendix:error_arithmetic}. This motivates the ``equation instruction''. Besides, we find out through statistics that the error overlap of code prompting and CoT prompting is very small, suggesting that the two prompting methods lead LLMs to think from different angles. (See Appendix~\ref{appendix:error_arithmetic} for more detail.) 

\subsection{Ensemble of CoT and code prompting}
\label{ensemble}

\begin{wrapfigure}[15]{r}{6cm}
\centering
\vspace{-5pt}
\begin{tabular}{cc}
\toprule
method  & GSM8K \\\midrule
few-shot   CoT  & 81.58 \\
few-shot code  & 79.68 \\\midrule
\begin{tabular}[c]{@{}c@{}}ensemble \\    (only CoT votes)\end{tabular}  & 87.49 \\\midrule
\begin{tabular}[c]{@{}c@{}}ensemble \\    (only code votes)\end{tabular} & 83.85 \\\midrule
\begin{tabular}[c]{@{}c@{}}ensemble \\    (both methods vote)\end{tabular}         & \textbf{87.95}\\\bottomrule
\end{tabular}
\setlength{\abovecaptionskip}{0.3cm}
\captionof{table}{Accuracy (\%) of ensemble methods on GSM8K.}
\label{table:ensemble}
\end{wrapfigure}
    
    Conceptually similar to~\citet{selfconsistency}, we consider the ensemble of CoT prompting and code prompting based on voting. For each question, if both prompting methods generate the same answer, we accept the answer as the final answer; otherwise, we set the temperature to 0.7 and ask the LLM to generate $n$ answers following each prompting method. Then we vote among the $2n$ answers to give the final answer. We also compare voting methods whose $2n$ answers are generated from the same prompting method. 
    As shown in Table~\ref{table:ensemble}, ensemble methods outperform the baselines significantly. Further, the ensemble of two prompting methods surpasses only code prompting votes or only CoT prompting votes as we expected. See Appendix~\ref{appendix:exp_arithmetic} for experiment details.

\subsection{Ablation study}
We add annotations in the exemplars of few-shot code prompting. To explore whether and where we should add the annotations, we conduct the following ablation experiments.

\begin{table}[t]
    \centering
    \begin{tabular}{cccccc}
    \toprule
    method & SingleEq & AddSub & MultiArith & SVAMP & GSM8K \\\midrule
    few-shot code w/o ann  & \textbf{97.64}    & 89.11  & 97.00      &  79.50 & 79.45\\\midrule
    few-shot code w/ ann at the end  & 97.44    & \textbf{89.87}  & \textbf{97.33}      &  79.60 & \textbf{80.21}\\\midrule
    few-shot code w/ ann at the beginning & 96.85    & \textbf{89.87}  & 95.33      & \textbf{79.90} & 77.71\\\bottomrule
    \end{tabular}
    \setlength{\abovecaptionskip}{0.3cm}
    \caption{Accuracy (\%) of methods of few-shot code prompting w/ annotations or w/o annotations added at different locations of each line of the code. ``Self-debugging'' is not used here.}
    \label{table:location}
\end{table}
    
\textbf{Do annotations matter?}
    We compare the code prompting with and without annotations. As shown in Table~\ref{table:location}, adding annotations generally raises the accuracy. 
    This demonstrates the benefits of adding annotations for few-shot code prompting. 
    Annotations written in natural language may assist the LLM to understand the exemplars better and generate better code.

\textbf{Does where we add annotations matter?} 
\label{location}
     We experiment on code prompting with annotations added either at the beginning or at the end of each line of code. As shown in Table~\ref{table:location}, in most cases, adding annotations at the end is a better choice. We suspect this is due to humans' habit of writing code. We tend to add annotations specifically describing each line of code at the end of each line more often than before the code, thus the training corpus contains more code samples with annotations at the end of each line. 

\textbf{Do annotations help harder or easier questions?}
    We experiment on GSM8K to find out where the performance gain of adding annotations comes from. We consider the number of words in the provided answer as a notion of difficulty. Figure~\ref{fig:difficulty} shows the performance gain of adding annotations to questions with different difficulty. The performance gain on the easiest questions and hardest questions is larger than that on the questions with medium difficulty.

\textbf{Self-debugging.}
Figure~\ref{fig:self-debug-ablation} shows the accuracy of zero-shot and few-shot code prompting with or without ``self-debugging'' on 5 arithmetic datasets. It is clear that the performance on harder benchmarks is more likely to benefit from ``self-debugging''. This is because more complex problems are more likely to trigger bugs in code. Meanwhile, few bugs are observed in SingleEq and MultiArith. 

\textbf{Irrelevant information.}
Through error analysis of AddSub in \S\ref{error:addsub}, we show that model performance may suffer from irrelevant information in questions. So we propose  ``irrelevant information'', introduced in~\S\ref{zero-shot code}. Table~\ref{table:irr} shows that despite the simplicity, it effectively improves the performance. Note that code prompting methods in the table are all equipped with ``self-debugging''.

\begin{minipage}[t]{1\textwidth}
\centering
\begin{minipage}[t]{0.375\textwidth}
\makeatletter\def\@captype{figure}
\centering
    \includegraphics[width=\textwidth]{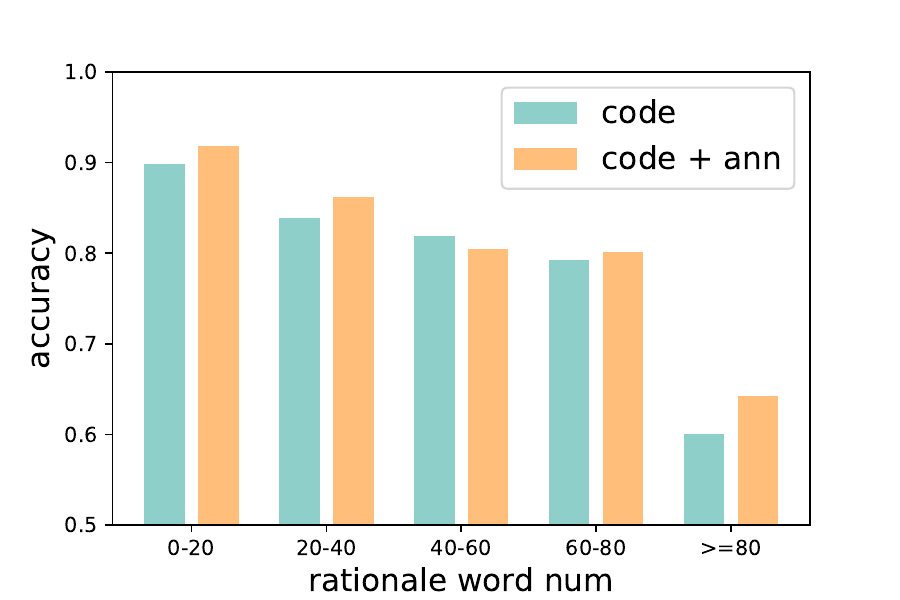}
    \vspace{-15pt}
    \caption{Accuracy of few-shot code prompting w/ or w/o annotations on questions with various difficulty.}
    \label{fig:difficulty}
\end{minipage}
\hspace{.05in}
\begin{minipage}[t]{0.60\textwidth}
\makeatletter\def\@captype{figure}
\centering
\includegraphics[width=\textwidth]{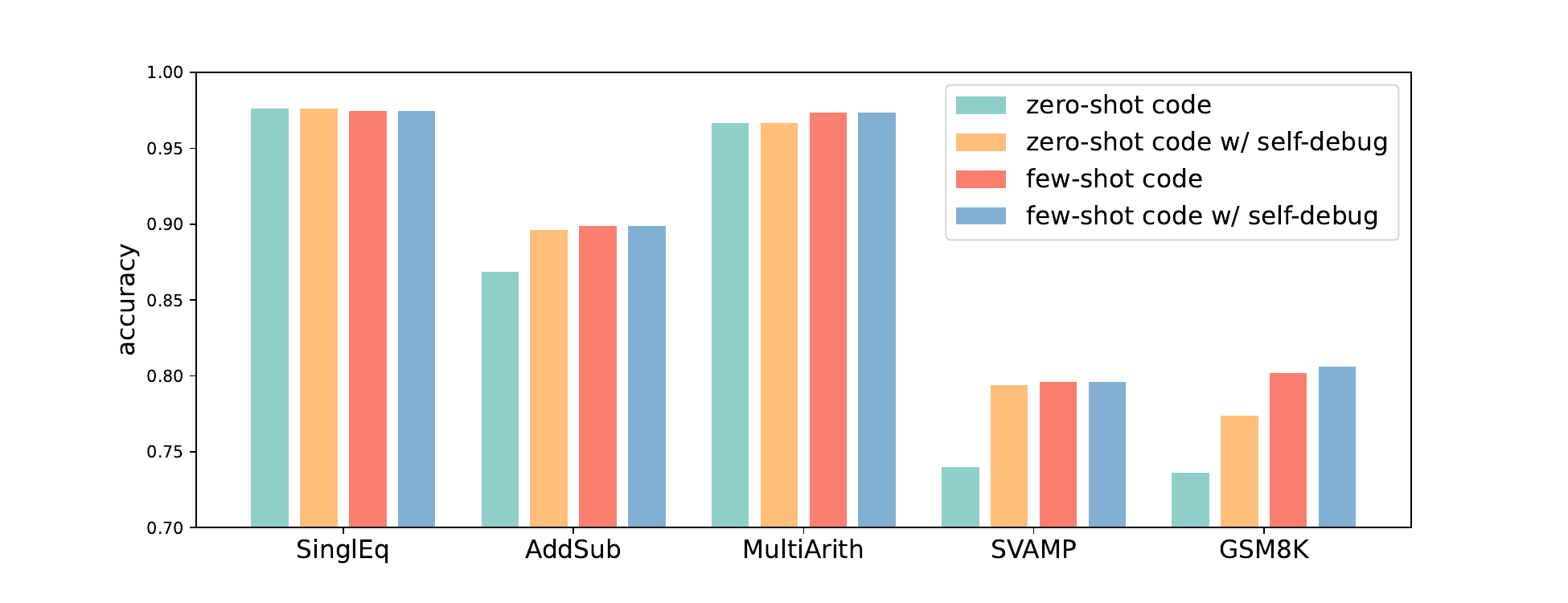}
    \vspace{-15pt}
    \caption{Accuracy of zero-shot and few-shot code prompting w/ or w/o ``self-debugging'' on arithmetic datasets.}
    \label{fig:self-debug-ablation}
\end{minipage}
\end{minipage}

\textbf{Equation instruction.}
As mentioned in~\S\ref{error:gsm8k}, it is challenging for LLMs to solve equations in the code. So we give an instruction on employing \texttt{sympy} to leverage Python to solve equations (see Appendix~\ref{appendix:exp_arithmetic}). Besides, we also experiment on directly asking the LLM to solve equations in the annotations. Table~\ref{table:equ} shows that the best choice is to teach the LLM to use certain Python packages to solve equations. Note that code prompting methods in the table all have ``self-debugging''.

\vspace{20pt}
\begin{minipage}[t]{\textwidth}
\centering
\begin{minipage}[t]{0.45\textwidth}
\makeatletter\def\@captype{table}
\centering
\begin{tabular}{rc}
    \toprule
    method  & AddSub \\\midrule
    zero-shot code & 89.62 \\
    zero-shot code$_{\text{+irr}}$ & \textbf{91.65} \\\midrule
    few-shot code & 89.87 \\
    few-shot code$_{\text{+irr}}$ & \textbf{91.39}
    \\\bottomrule
\end{tabular}
\caption{Accuracy (\%) of code prompting w/ or w/o ``irrelevant information'' on AddSub.}
\label{table:irr}

\end{minipage}
\hspace{.15in}
\begin{minipage}[t]{0.4\textwidth}
\makeatletter\def\@captype{table}
\centering
\begin{tabular}{rc}
    \toprule
    method  & GSM8K \\\midrule
    zero-shot code & 77.40 \\
    zero-shot code$_{\text{+equ(ann)}}$ & 78.09 \\
    zero-shot code$_{\text{+equ(sympy)}}$ & \textbf{78.92}\\
    \bottomrule
\end{tabular}
\caption{Accuracy (\%) of code prompting w/ or w/o ``equation instruction'' on GSM8K. }
\label{table:equ}
\end{minipage}
\end{minipage}

\section{Discussion}
\label{discussion}

\begin{wrapfigure}[23]{r}{6cm}
    \centering
    \vspace{-20pt}
    \includegraphics[width=\textwidth]{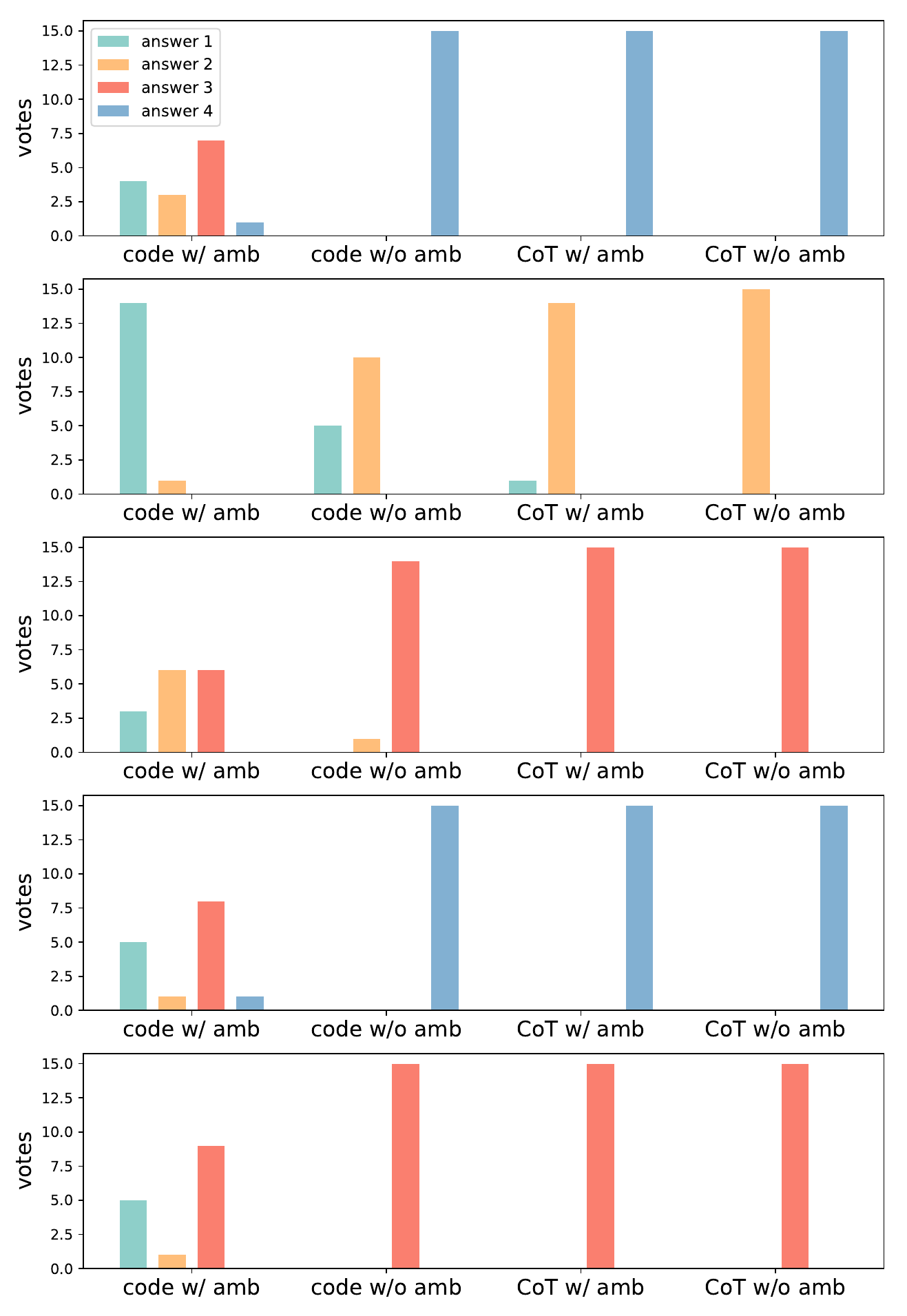}
    \caption{Answer distribution on questions w/ and w/o ambiguity.}
    \label{fig:disambiguity}
    \vspace{-10pt}
\end{wrapfigure}

\textbf{Disambiguation.} 
Following the thought mentioned in~\S\ref{error analysis}, we find out through experiments that code prompting has the potential to discover ambiguity in a question. Here we use 5 cases where the questions present ambiguities and we can manually fix them. We test few-shot code prompting and CoT prompting on both questions with and without ambiguity. For each question, we generate 15 answers by code prompting and CoT prompting respectively with the temperature of the LLM set to 0.7. Figure~\ref{fig:disambiguity} shows the histogram of answers. It is evident that ambiguity disturbs code prompting, while CoT prompting is less sensitive to ambiguity. We may leverage this feature to detect ambiguity in questions. 


\textbf{Zero-shot methods vs. few-shot methods}
As Large Language Models develop in size and human-alignment, the gap between zero-shot and few-shot prompting methods seems to narrow~\citep{zhong2023agieval}. We have shown in the paper that zero-shot code prompting is competitive with current few-shot methods, including few-shot CoT~\citep{few-shot_cot} and PAL~\citep{pal}. However, the reasons behind this are still left to be explored.

\section{Conclusion}
We introduce code prompting, a neural-symbolic prompting method with both zero-shot and few-shot versions. We conduct experiments on 7 popular benchmarks involving both symbolic and arithmetic reasoning and show that both zero-shot and few-shot code prompting enhance complex reasoning in LLMs. Extensive experiments and analyses verify the effectiveness of our self-debugging, annotation, equation instruction and irrelevant information techniques, and present many insights. 



\bibliographystyle{abbrvnat}  
\bibliography{references}  

\clearpage
\appendix
\section{Datasets}
\label{appendix:data}
\subsection{Symbolic reasoning}
\label{appendix:data_symbolic}
\begin{table}[H]
\begin{tabularx}{\textwidth}{
>{\hsize=.8\hsize\linewidth=\hsize}X
>{\hsize=.8\hsize\linewidth=\hsize}>{\centering\arraybackslash}X
>{\hsize=.8\hsize\linewidth=\hsize}>{\centering\arraybackslash}X
>{\hsize=1.6\hsize\linewidth=\hsize}X
}
\toprule
dataset & \multicolumn{2}{c}{\#samples} & example problem    \\\midrule
\multirow{3}{*}{last letter   concatenation} & \multirow{3}{*}{1,500} & 500 (4 words)  & "fully, drug, gut, agreement"  \\
& & 500 (8 words)  & "urge, participate, strategic, rise, win, through, now, within" \\
& & 500 (12 words) & "elementary, consensus, plaza, message, telescope, accessible, another, transport, bubble, bizarre,   adviser, cow"  \\\midrule
\multirow{3}{*}{coin flip}  & \multirow{3}{*}{1,500} & 500 (3 flips)  & A coin is heads up. Taylor doesn't flip the coin. Harmon doesn't flip the coin. Dejesus doesn't flip the coin. Is the coin still heads up? \\
& & 500 (4 flips)  & A coin is heads up. Nichols flips the coin. Mcbride flips the coin. Mathis doesn't flip the coin. Ruiz   flips the coin. Is the coin still heads up? \\
& & 500 (5 flips)  & A coin is heads up. Weber flips the coin. Contreras flips the coin. Carter flips the coin. Moran flips the   coin. Colon doesn't flip the coin. Is the coin still heads up? 
\\\bottomrule
\end{tabularx}
\end{table}
\subsection{Arithmetic reasoning}
\label{appendix:data_arithmetic}
\begin{table}[H]
\begin{tabularx}{\textwidth}{
>{\hsize=.6\hsize\linewidth=\hsize}X
>{\hsize=.6\hsize\linewidth=\hsize}X
>{\hsize=1.8\hsize\linewidth=\hsize}X 
}
\toprule
dataset    & \#samples & example problem  \\\midrule
SingleEq   & 508       & Joan found 70 seashells on the   beach. she gave Sam some of her seashells. She has 27 seashell left. How many   seashells did she give to Sam ?   \\\midrule
AddSub     & 395       & There are 22 walnut trees   currently in the park. Park workers will plant walnut trees today. When the   workers are finished there will be 55 walnut trees in the park. How many   walnut trees did the workers plant today?     \\\midrule
MultiArith & 600       & For Halloween Debby and her   sister combined the candy they received. Debby had 32 pieces of candy while   her sister had 42. If they ate 35 pieces the first night, how many pieces do   they have left?                          \\\midrule
SVAMP      & 1,000     & Julia played tag with 18 kids on monday. She played tag with 10 kids on tuesday. How many more kids did she play with on monday than on tuesday ?  \\\midrule
GSM8K      & 1,319     & Janet’s ducks lay 16 eggs per   day. She eats three for breakfast every morning and bakes muffins for her   friends every day with four. She sells the remainder at the farmers' market   daily for \$2 per fresh duck egg. How much in dollars does she make every day   at the farmers' market?
\\\bottomrule
\end{tabularx}
\end{table}

\section{Experiment details for symbolic reasoning}
\label{appendix:exp_symbolic}
\subsection{Last letter concatenation}
Here we show correct and incorrect examples for all the prompting methods we have mentioned in the paper on the task of last letter concatenation.

\textbf{Zero-shot standard}

\begin{table}[H]
\begin{tabular}{l}
\toprule
\begin{tabular}[c]{@{}l@{}}Q:   Concatenate the last letters of the given words: "imagine,admire,assume,equally"\\ A:   e, e, e, y. Therefore, the answer is "eeey“   {\color{green!60!black}\textbf{(correct)}}\end{tabular}           \\\midrule
\begin{tabular}[c]{@{}l@{}}Q:   Concatenate the last letters of the given words: uh,employer,intelligent,straight"\\ A:   hrtt.   Therefore, the answer is "hrtt". {\color{green!60!black}\textbf{(correct)}}\end{tabular}            \\\midrule
\begin{tabular}[c]{@{}l@{}}Q:   Concatenate the last letters of the given words: "motion,instant,act,chapter"\\A:   n,t,t,r.   Therefore, the answer is ``nttr''. {\color{green!60!black}\textbf{(correct)}}\end{tabular}              \\\midrule
\begin{tabular}[c]{@{}l@{}}Q:   Concatenate the last letters of the given words: "fully,drug,gut,agreement"\\ A:   lygdnt.   Therefore, the answer is ``lygdnt'' {\color{red!60!black}\textbf{(incorrect)}}\end{tabular}              \\\midrule
\begin{tabular}[c]{@{}l@{}}Q:   Concatenate the last letters of the given words: "evil,secret,expansion,experience"\\A:   lnts.   Therefore, the answer is lnts. {\color{red!60!black}\textbf{(incorrect)}}\end{tabular}           \\\midrule
\begin{tabular}[c]{@{}l@{}}Q:   Concatenate the last letters of the given words: "consist,wear,concede,considerable"\\ A:   t-r-e-e-l-e. Therefore,   the answer is "treele". {\color{red!60!black}\textbf{(incorrect)}}
\end{tabular}
\\\bottomrule
\end{tabular}
\caption{Examples of correct and incorrect results by zero-shot standard prompting on last letter concatenation.}
\label{llt_zero-shot_standard}
\end{table}

\clearpage
\textbf{Zero-shot CoT}

\begin{table}[H]
\begin{tabular}{l}
\toprule
\begin{tabular}[c]{@{}l@{}}Q:   Concatenate the last letters of the given words: "fully,drug,gut,agreement"\\ A:   Let's think step by step.\\1.   The last letter of "fully" is "y".\\ 2.   The last letter of "drug" is "g".\\ 3.   The last letter of "gut" is "t".\\  4.   The last letter of "agreement" is "t".\\  5.   Concatenating these letters together, we get "ygtt".Therefore,   the answer is "ygtt".  {\color{green!60!black}\textbf{(correct)}}\end{tabular}          \\\midrule
\begin{tabular}[c]{@{}l@{}}Q:   Concatenate the last letters of the given words: "imagine,admire,assume,equally" \\ A:   Let's think step by step.1. The last letter of "imagine" is   "e".\\ 2.   The last letter of "admire" is "e".\\ 3.   The last letter of "assume" is "e".\\ 4.   The last letter of "equally" is "y".\\  5.   Concatenating these letters gives us "eeeY".Therefore,   the answer is "eeeY".  {\color{green!60!black}\textbf{(correct)}}\end{tabular}                 \\\midrule
\begin{tabular}[c]{@{}l@{}}Q:   Concatenate the last letters of the given words: "consist,wear,concede,considerable"\\  A:   Let's think step by step.1. The last letter of "consist" is   "t".\\ 2.   The last letter of "wear" is "r".\\  3.   The last letter of "concede" is "e".\\ 4.   The last letter of "considerable" is "e".\\ 5.   Concatenating these letters together, we get "tree".Therefore,   the answer is "tree".{\color{green!60!black}\textbf{(correct)}}\end{tabular}   \\\midrule
\begin{tabular}[c]{@{}l@{}}Q:   Concatenate the last letters of the given words: "evil,secret,expansion,experience"\\  A:   Let's think step by step.1. The last letter of "evil" is   "l".\\  2.   The last letter of "secret" is "t".\\ 3.   The last letter of "expansion" is "n".\\4.   The last letter of "experience" is "e".\\  5.   Concatenating these letters together, we get "lten". Therefore, the answer is "lten". {\color{red!60!black}\textbf{(incorrect)}}\end{tabular} \\\midrule
\begin{tabular}[c]{@{}l@{}}Q:   Concatenate the last letters of the given words: "weapon,reasonable,thin,loyalty"\\ A:   Let's think step by step.1. The last letter of "weapon" is   "n".\\  2.   The last letter of "reasonable" is "e".\\  3.   The last letter of "thin" is "n".\\  4.   The last letter of "loyalty" is "y".\\  5.   Concatenating these letters together, we get "ney".Therefore,   the answer is "ney". {\color{red!60!black}\textbf{(incorrect)}}\end{tabular}  \\\midrule
\begin{tabular}[c]{@{}l@{}}Q:   Concatenate the last letters of the given words: "motion,instant,act,chapter"\\ A:   Let's think step by step.1. The last letter of "motion" is   "n".\\2.   The last letter of "instant" is "t".\\ 3.   The last letter of "act" is "t".\\  4.   The last letter of "chapter" is "r".\\ 5.   Concatenating these letters together, we get "ntr".Therefore, the answer is "ntr". {\color{red!60!black}\textbf{(incorrect)}}\end{tabular}   \\\bottomrule           
\end{tabular}
\caption{Examples of correct and incorrect results by zero-shot CoT prompting on last letter concatenation.}
\label{llt_zero-shot_standard}
\end{table}

\clearpage
\textbf{Few-shot CoT}

\begin{table}[H]
\begin{tabular}{l}
\toprule
\begin{tabularx}{\textwidth}{X}
Q: "think, machine"\\ A: The last letter of "think"   is "k". The last letter of "machine" is "e".   Concatenating "k", "e" leads to "ke". So, "think, machine"   outputs "ke".\\ \\ Q: "learning, reasoning, generalization"\\ A: The last letter of "learning"is "g". The last letter   of "reasoning" is "g". The last letter of   "generalization" is "n". Concatenating "g",   "g", "n" leads to "ggn". So, "learning, reasoning,   generalization" outputs "ggn".\end{tabularx}
\\\bottomrule
\end{tabular}
\caption{Exemplars for the full prompt of few-shot CoT prompting on last letter concatenation. We follow~\citet{ltm} to construct the exemplars.}
\end{table}

\begin{table}[H]
\begin{tabular}{l}
\toprule
\begin{tabularx}{\textwidth}{X}Q:   Concatenate the last letters of the given words: "fully,drug,gut,agreement"\\ A:   The last letter of "fully" is "y". The last letter of   "drug" is "g". The last letter of "gut" is   "t". The last letter of "agreement" is "t".   Concatenating "y", "g", "t", "t"   leads to "ygtt".   So, "fully,drug,gut,agreement"   outputs "ygtt".{\color{green!60!black}\textbf{(correct)}}\end{tabularx}                         \\\midrule
\begin{tabularx}{\textwidth}{X}Q:   Concatenate the last letters of the given words: "imagine,admire,assume,equally"\\ A:   The last letter of "imagine" is "e". The last letter of   "admire" is "e". The last letter of "assume" is   "e". The last letter of "equally" is "y".   Concatenating "e", "e", "e", "y"   leads to "eeey".   So, "imagine, admire, assume, equally" outputs "eeey".{\color{green!60!black}\textbf{(correct)}}\end{tabularx}         \\\midrule
\begin{tabularx}{\textwidth}{X}Q:   Concatenate the last letters of the given words: "evil,secret,expansion,experience"\\ A:   The last letter of "evil" is "l". The last letter of   "secret" is "t". The last letter of "expansion"   is "n". The last letter of "experience" is "e".   Concatenating "l", "t", "n", "e"   leads to "ltne".   So, "evil,secret,expansion,experience"   outputs "ltne".{\color{green!60!black}\textbf{(correct)}}\end{tabularx} \\\midrule
\begin{tabularx}{\textwidth}{X}Q:   "vacation,diverse,argument,speculation"\\ A:   The last letter of "vacation" is "n". The last letter of   "diverse" is "e". The last letter of "argument"   is "t". The last letter of "speculation" is   "n". Concatenating "n", "e", "t",   "n" leads to "nent".   So, "vacation, diverse, argument, speculation" outputs "nent".  {\color{red!60!black}\textbf{(incorrect)}}\end{tabularx} \\\midrule
\begin{tabularx}{\textwidth}{X}Q:   "Christianity,arm,invention,dog"\\ A:   There is no common letter at the end of all four words, so there is no output   for "Christianity, arm, invention, dog".{\color{red!60!black}\textbf{(incorrect)}}\end{tabularx}  \\\midrule
\begin{tabularx}{\textwidth}{X}Q:   "German,establishment,accurate,initially"  \\ A:   The last letter of "German" is "n". The last letter of   "establishment" is "t". The last letter of   "accurate" is "e". The last letter of   "initially" is "y". Concatenating "n",   "t", "e", "y" leads to "nty".   So, "German, establishment, accurate, initially" outputs "nty".  {\color{red!60!black}\textbf{(incorrect)}}\end{tabularx}
\\\bottomrule
\end{tabular}
\caption{Examples of correct and incorrect results by few-shot CoT prompting on last letter concatenation.}
\end{table}

\clearpage
\textbf{Zero-shot code}

\begin{table}[H]
\begin{tabular}{l}
\toprule
Generate python code to concatenate the last letters of the given words.   \\ \\

Here's the Python code to concatenate the last letters of the given words: \\
\Verb|`|\Verb|`|\Verb|`|                                                                       \\
words = {[}"apple", "banana", "cherry", "date", "elderberry"{]}            \\
result = ""                                                                \\
for word in words:                                                         \\
result += word{[}-1{]}                                                     \\
print(result)                                                              \\
\Verb|`|\Verb|`|\Verb|`|  \\\bottomrule
\end{tabular}
\caption{Code generated in the first stage of code prompting for last letter concatenation.}
\end{table}

\begin{table}[H]
\begin{tabular}{l}
\toprule
\begin{tabular}[c]{@{}l@{}}Q:   "fully,drug,gut,agreement"\\ A:\\ Let's   think step by step. Print all the middle variables.\\ \Verb|`|\Verb|`|\Verb|`|\\ words   = {[}"fully", "drug", "gut",   "agreement"{]}\\ result   = ""\\ for   word in words:\\     last\_letter   = word{[}-1{]}\\     result += last\_letter\\     print(f"Word:   \{word\}, Last Letter: \{last\_letter\},   Result: \{result\}")\\ print(result)\\ \Verb|`|\Verb|`|\Verb|`|\\ Output:\\ \Verb|`|\Verb|`|\Verb|`|\\ Word:   fully, Last Letter: y, Result: y\\ Word:   drug, Last Letter: g, Result: yg\\ Word:   gut, Last Letter: t, Result: ygt\\ Word:   agreement, Last Letter: t, Result: ygtt\\ ygtt\\ \Verb|`|\Verb|`|\Verb|`|\\ Therefore,   the answer is "ygtt". {\color{green!60!black}\textbf{(correct)}}
\end{tabular}                  \\\midrule
\begin{tabular}[c]{@{}l@{}}Q:   "feed,characteristic,spray,anymore"\\ A:\\ Let's   think step by step. Print all the middle variables.\\ \Verb|`|\Verb|`|\Verb|`|\\ words   = {[}"feed", "characteristic", "spray",   "anymore"{]}\\ result   = ""\\ for   word in words:\\     last\_letter   = word{[}-1{]}\\     result += last\_letter\\     print(f"Word:   \{word\}, Last Letter: \{last\_letter\},   Result: \{result\}")\\ print(result)\\ \Verb|`|\Verb|`|\Verb|`|\\ Output:\\ \Verb|`|\Verb|`|\Verb|`|\\ Word:   feed, Last Letter: d, Result: d\\ Word:   characteristic, Last Letter: c, Result: dc\\ Word:   spray, Last Letter: y, Result: dcy\\ Word:   anymore, Last Letter: e, Result: dbye\\ dbye\\ \Verb|`|\Verb|`|\Verb|`|\\ Therefore,   the answer is dbye. {\color{red!60!black}\textbf{(incorrect)}}
\end{tabular}\\\bottomrule
\end{tabular}
\caption{Examples of correct and incorrect results (in the second stage) by zero-shot code prompting on last letter concatenation.}
\end{table}

\clearpage
\subsection{Coin flip}
Here we show correct and incorrect examples for all the prompting methods we have mentioned in the paper on the task of coin flip.

\textbf{Zero-shot standard}

\begin{table}[H]
\begin{tabular}{l}
\toprule
\begin{tabularx}{\textwidth}{X}Q:   A coin is heads up. Taylor doesn't flip the coin. Harmon doesn't flip the   coin. Dejesus   doesn't flip the coin. Is the coin still heads up? Note that "flip"   here means "reverse".  \\ A:   Yes, the coin is still heads up. Therefore, the answer (Yes or No) is Yes. 
 {\color{green!60!black}\textbf{(correct)}}\end{tabularx}\\\midrule
\begin{tabularx}{\textwidth}{X}Q:   A coin is heads up. Tyler doesn't flip the coin. Gates doesn't flip the coin.   Weber doesn't flip the coin. Is the coin still heads up? Note that   "flip" here means "reverse".\\    \\ A:   Yes, the coin is still heads up. Therefore, the answer (Yes or No) is Yes. {\color{green!60!black}\textbf{(correct)}}\end{tabularx}\\\midrule
\begin{tabularx}{\textwidth}{X}Q:   A coin is heads up. Kline doesn't flip the coin. Underwood doesn't flip the   coin. Yu doesn't flip the coin. Is the coin still heads up? Note that   "flip" here means "reverse".\\    \\ A:   Yes, the coin is still heads up. The fact that no one flipped the coin does   not change its initial state. Therefore, the answer (Yes or No) is Yes. {\color{green!60!black}\textbf{(correct)}}\end{tabularx} \\\midrule
\begin{tabularx}{\textwidth}{X}Q:   A coin is heads up. Valencia doesn't flip the coin. Ross flips the coin.   Walter doesn't flip the coin. Is the coin still heads up? Note that   "flip" here means "reverse".\\    \\ A:   Yes, the coin is still heads up. The actions of Valencia, Ross, and Walter do   not affect the initial state of the coin. Therefore, the answer (Yes or No)   is Yes. {\color{red!60!black}\textbf{(incorrect)}}\end{tabularx}                                                                                         \\\midrule
\begin{tabularx}{\textwidth}{X}Q:   A coin is heads up. Valentine flips the coin. Beck flips the coin. Molina   doesn't flip the coin. Is the coin still heads up? Note that "flip"   here means "reverse".\\    \\ A:   No, we don't know the result of Valentine and Beck's flips, so we can't   assume the coin is still heads up. Therefore, the answer (Yes or No) is   unknown. {\color{red!60!black}\textbf{(incorrect)}}\end{tabularx}                                                                                                        \\\midrule
\begin{tabularx}{\textwidth}{X}Q:   A coin is heads up. Stephens flips the coin. Cooper doesn't flip the coin.   Olson flips the coin. Is the coin still heads up? Note that "flip"   here means "reverse".\\    \\ A:   It depends on the outcome of Olson's flip. If Olson flips the coin and it   lands heads up, then the coin is still heads up. If Olson flips the coin and   it lands tails up, then the coin is no longer heads up.Therefore,   the answer (Yes or No) is uncertain. {\color{red!60!black}\textbf{(incorrect)}}\end{tabularx}\\\bottomrule
\end{tabular}
\caption{Examples of correct and incorrect results by zero-shot standard prompting on coin flip.}
\end{table}

\clearpage
\textbf{Zero-shot CoT}
\begin{table}[H]
\begin{tabular}{l}
\toprule
\begin{tabularx}{\textwidth}{X}Q:   A coin is heads up. Taylor doesn't flip the coin. Harmon doesn't flip the   coin. Dejesus doesn't flip the coin. Is the coin still heads up? Note that   "flip" here means "reverse".\\ A:   Let's think step by step. Initially, the coin is heads up. None of the three   people mentioned in the question flip the coin, which means they don't   reverse it. Therefore, the coin remains in its initial state, which is heads   up. So, yes, the coin is still heads up. Therefore, the answer (Yes or No) is   Yes. {\color{green!60!black}\textbf{(correct)}}\end{tabularx} \\\midrule
\begin{tabularx}{\textwidth}{X}Q:   A coin is heads up. Valencia doesn't flip the coin. Ross flips the coin.   Walter doesn't flip the coin. Is the coin still heads up? Note that   "flip" here means "reverse".\\ A:   Let's think step by step.\\ 1.   The coin is heads up.\\ 2.   Valencia doesn't flip the coin, so the coin remains heads up.\\ 3.   Ross flips the coin, which means the coin is now tails up.\\ 4.   Walter doesn't flip the coin, so the coin remains tails up. \\ Therefore,   the answer is no, the coin is not still heads up. It is now tails   up. Therefore, the answer (Yes or No) is No. {\color{green!60!black}\textbf{(correct)}}\end{tabularx}\\\midrule
\begin{tabularx}{\textwidth}{X}Q:   A coin is heads up. Valentine flips the coin. Beck flips the coin. Molina   doesn't flip the coin. Is the coin still heads up? Note that "flip"   here means "reverse".\\ A:   Let's think step by step.1. The coin is heads up.\\ 2.   Valentine flips the coin, which means it is now tails up.\\ 3.   Beck flips the coin, which means it is now heads up again.\\ 4.   Molina doesn't flip the coin, so it remains heads up.\\ Therefore,   the coin is still heads up. Therefore, the answer (Yes or No) is Yes. {\color{green!60!black}\textbf{(correct)}}\end{tabularx}  \\\midrule
\begin{tabularx}{\textwidth}{X}Q:   A coin is heads up. Medrano doesn't flip the coin. Cannon doesn't flip the   coin. Hutchinson flips the coin. Is the coin still heads up? Note that   "flip" here means "reverse".\\ A:   Let's think step by step.\\ 1.   The coin is heads up.\\ 2.   Medrano doesn't flip the coin, so it remains heads up.\\ 3.   Cannon doesn't flip the coin, so it still remains heads up.\\ 4.   Hutchinson flips the coin. We don't know the result of the flip, so we can't   say for sure whether the coin is still heads up or not.\\ Therefore,   we can't determine whether the coin is still heads up or not after Hutchinson   flips it. Therefore, the answer (Yes or No) is indeterminate. {\color{red!60!black}\textbf{(incorrect)}}\end{tabularx}       \\\midrule
\begin{tabularx}{\textwidth}{X}Q:   A coin is heads up. Farmer flips the coin. Chandler doesn't flip the coin.   Lozano flips the coin. Is the coin still heads up? Note that "flip"   here means "reverse".\\ A:   Let's think step by step.\\ 1.   The coin is heads up.\\ 2.   Farmer flips the coin, which means he reverses it. So now the coin is tails   up.\\ 3.   Chandler doesn't flip the coin, so it remains tails up.\\ 4.   Lozano flips the coin, which means he reverses it again. So now the coin is   heads up again.\\ Therefore,   the coin is not still heads up, but it has gone through a series of flips and   is now heads up again. Therefore, the answer (Yes or No) is No. {\color{red!60!black}\textbf{(incorrect)}}\end{tabularx}                                 \\\midrule
\begin{tabularx}{\textwidth}{X}Q:   A coin is heads up. Christian doesn't flip the coin. Lewis doesn't flip the   coin. Becker flips the coin. Is the coin still heads up? Note that   "flip" here means "reverse".\\ A:   Let's think step by step.\\ 1.   The coin is heads up.\\ 2.   Christian doesn't flip the coin, so it remains heads up.\\ 3.   Lewis doesn't flip the coin, so it still remains heads up.\\ 4.   Becker flips the coin. We don't know the result of the flip, so we can't say   for sure whether the coin is still heads up or not.\\ Therefore,   we can't determine whether the coin is still heads up or not without knowing   the result of Becker's flip. Therefore, the answer (Yes or No) is unknown. {\color{red!60!black}\textbf{(incorrect)}}\end{tabularx}\\\bottomrule
\end{tabular}
\caption{Examples of correct and incorrect results by zero-shot CoT prompting on coin flip.}
\end{table}

\clearpage
\textbf{Few-shot CoT}
\begin{table}[H]
\begin{tabularx}{\textwidth}{X}\toprule
Q: A coin is heads up. Ka flips the coin. Sherrie flips the coin. Is the coin still heads up? Note that "flip" here means "reverse".                                                                                  \\
A: The coin was flipped by Ka and Sherrie. So the coin was flipped 2 times, which is an even number. The coin started heads up, so after an even number of flips, it will still be heads up. So the answer is yes.    \\ \\
Q: A coin is heads up. Jamey flips the coin. Teressa flips the coin. Is the coin still heads up? Note that "flip" here means "reverse".                                                                               \\
A: The coin was flipped by Jamey and Teressa. So the coin was flipped 2 times, which is an even number. The coin started heads up, so after an even number of flips, it will still be heads up. So the answer is yes. \\ \\
Q: A coin is heads up. Maybelle flips the coin. Shalonda does not flip the coin. Is the coin still heads up? Note that "flip" here means "reverse".                                                                   \\
A: The coin was flipped by Maybelle. So the coin was flipped 1 time, which is an odd number. The coin started heads up, so after an odd number of flips, it will be tails up. So the answer is no.                    \\ \\
Q: A coin is heads up. Millicent does not flip the coin. Conception flips the coin. Is the coin still heads up? Note that "flip" here means "reverse".                                                                \\
A: The coin was flipped by Conception. So the coin was flipped 1 time, which is an odd number. The coin started heads up, so after an odd number of flips, it will be tails up. So the answer is no.                  \\ \\
Q: A coin is heads up. Sal flips the coin. Raymond does not flip the coin. Is the coin still heads up? Note that "flip" here means "reverse".                                                                         \\
A: The coin was flipped by Sal. So the coin was flipped 1 time, which is an odd number. The coin started heads up, so after an odd number of flips, it will be tails up. So the answer is no.                         \\ \\
Q: A coin is heads up. Conception flips the coin. Kristian does not flip the coin. Is the coin still heads up? Note that "flip" here means "reverse".                                                                 \\
A: The coin was flipped by Conception. So the coin was flipped 1 time, which is an odd number. The coin started heads up, so after an odd number of flips, it will be tails up. So the answer is no.                  \\ \\
Q: A coin is heads up. Inga does not flip the coin. Elanor does not flip the coin. Is the coin still heads up? Note that "flip" here means "reverse".                                                                 \\
A: The coin was flipped by no one. So the coin was flipped 0 times. The coin started heads up, and it was not flipped, so it is still heads up. So the answer is yes.                                                 \\ \\
Q: A coin is heads up. Ryan flips the coin. Shaunda flips the coin. Is the coin still heads up? Note that "flip" here means "reverse".                                                                                \\
A: The coin was flipped by Ryan and Shaunda. So the coin was flipped 2 times, which is an even number. The coin started heads up, so after an even number of flips, it will still be heads up. So the answer is yes. \\\bottomrule
\end{tabularx}
\caption{Exemplars for the full prompt of few-shot CoT prompting on coin flip. We follow~\citet{few-shot_cot} to construct the exemplars.}
\end{table}

\begin{table}[H]
\begin{tabular}{l}\toprule
\begin{tabularx}{\textwidth}{X}Q:   A coin is heads up. Taylor doesn't flip the coin. Harmon doesn't flip the   coin. Dejesus   doesn't flip the coin. Is the coin still heads up? Note that "flip"   here means "reverse".\\  A:   The coin was not flipped by anyone. So the coin was flipped 0 times. The coin   started heads up, and it was not flipped, so it is still heads up. So the   answer is yes. {\color{green!60!black}\textbf{(correct)}}
\end{tabularx}                                                                                                 \\\midrule
\begin{tabularx}{\textwidth}{X}Q:   A coin is heads up. Valencia doesn't flip the coin. Ross flips the coin.   Walter doesn't flip the coin. Is the coin still heads up? Note that   "flip" here means "reverse".\\  A:   The coin was flipped by Ross. So the coin was flipped 1 time, which is an odd   number. The coin started heads up, so after an odd number of flips, it will   be tails up. So the answer is no. {\color{green!60!black}\textbf{(correct)}}
\end{tabularx}                                                                                      \\\midrule
\begin{tabularx}{\textwidth}{X}Q:   A coin is heads up. Valentine flips the coin. Beck flips the coin. Molina   doesn't flip the coin. Is the coin still heads up? Note that "flip"   here means "reverse".\\  A:   The coin was flipped by Valentine and Beck. So the coin was flipped 2 times,   which is an even number. The coin started heads up, so after an even number   of flips, it will still be heads up. Molina did not flip the coin, so it does   not affect the outcome. So the answer is yes. {\color{green!60!black}\textbf{(correct)}}
\end{tabularx} \\\midrule
\begin{tabularx}{\textwidth}{X}Q:   A coin is heads up. Park flips the coin. Cooper doesn't flip the coin. Wright   doesn't flip the coin. Is the coin still heads up?\\  A:   The coin was flipped by Park. So the coin was flipped 1 time, which is an odd   number. The coin started heads up, so after an odd number of flips, it will   be tails up. However, since Cooper and Wright did not flip the coin, we do   not know if it was flipped again. So the answer is uncertain. {\color{red!60!black}\textbf{(incorrect)}}
\end{tabularx}\\\bottomrule                     
\end{tabular}
\caption{Examples of correct and incorrect results by few-shot CoT prompting on coin flip.}
\end{table}

\textbf{Zero-shot code}
\begin{table}[H]
\begin{tabularx}{\textwidth}{X}
\toprule
A coin is heads up , there are some people, each one flipped or didn't flip the coin. Generate python code to determine whether the coin is still heads up.                                                   \\
Note that "flip" here means "reverse".               \\
Assuming that the initial state of the coin is represented by a boolean variable called "heads\_up" and the list of people's actions is represented by a list of boolean values called "flips", the following Python code can be used to determine whether the coin is still heads up: \\
\Verb|`|\Verb|`|\Verb|`|                                                 \\
heads\_up = True  \# initial state of the coin       \\
flips = {[}True, False, True, True, False{]} \# list of people's actions                                  \\
for flip in flips:                                   \\
if flip:                                             \\
heads\_up = not heads\_up                            \\
\# if the person didn't flip the coin, do nothing    \\
if heads\_up: \\
print("The coin is still heads up.")                 \\
else:                                                \\
print("The coin is now tails up.")                   \\
\Verb|`|\Verb|`|\Verb|`|    \\\bottomrule                                             
\end{tabularx}
\caption{Code generated in the first stage of code prompting for coin flip.}
\end{table}

\begin{table}[H]
\begin{tabular}{l}
    \toprule
\begin{tabularx}{\textwidth}{X}Q:   A coin is heads up. Taylor doesn't flip the coin. Harmon doesn't flip the   coin. Dejesus   doesn't flip the coin. Is the coin still heads up? \\ Let's   think step by step. Print all the middle variables. \\ A:Yes,   the coin is still heads up.  \\ To   verify this, we can use the same Python code as before, but with a different   list of flips: \\ \Verb|`|\Verb|`|\Verb|`|\\ heads\_up   = True  \# initial state of the coin \\ flips   = {[}False, False, False{]}  \# list of   people's actions \\ for   flip in flips: \\     if flip: \\         heads\_up   = not heads\_up  \# reverse the state of the coin \\     \# if the person didn't flip the coin, do   nothing \\     print(heads\_up)  \# print the current state of the coin \\ if   heads\_up: \\     print("The coin is still heads   up.") \\ else: \\     print("The coin is now tails   up.") \\ \Verb|`|\Verb|`|\Verb|`|\\  \\ When   we run this code, we can see that the current state of the coin remains True   (heads up) throughout the loop, and the final message confirms that the coin   is still heads up: \\ \Verb|`|\Verb|`|\Verb|`|\\ True \\ True \\ True \\ The   coin is still heads up. \\ \Verb|`|\Verb|`|\Verb|`|\\ Therefore,   the answer (Yes or No) is Yes. {\color{green!60!black}\textbf{(correct)}}
\end{tabularx} \\\midrule
\begin{tabularx}{\textwidth}{X}Q:   A coin is heads up. Valencia doesn't flip the coin. Ross flips the coin.   Walter doesn't flip the coin. Is the coin still heads up? \\ Let's   think step by step. Print all the middle variables. \\ A:Using   the same code as before, we can modify the list of flips to match the   scenario described in the question: \\ \Verb|`|\Verb|`|\Verb|`|\\ heads\_up   = True  \# initial state of the coin \\ flips   = {[}False, True, False{]}  \# list of   people's actions \\ for   flip in flips: \\     if flip: \\         heads\_up   = not heads\_up  \# reverse the state of the coin \\     \# if the person didn't flip the coin, do   nothing \\     print(heads\_up) \\ if   heads\_up: \\     print("The coin is still heads   up.") \\ else: \\     print("The coin is now tails   up.") \\ \Verb|`|\Verb|`|\Verb|`|\\  \\ When   we run this code, we get the following output: \\ \Verb|`|\Verb|`|\Verb|`|\\ True \\ False \\ True \\ The   coin is still heads up. \\ \Verb|`|\Verb|`|\Verb|`|\\ This   means that the coin is still heads up after Ross flips it, since the final   state of the coin is still heads up.Therefore,   the answer (Yes or No) is Yes, the coin is still heads up. {\color{red!60!black}\textbf{(incorrect)}}
\end{tabularx}    \\\bottomrule                                        
\end{tabular}
\caption{Examples of correct and incorrect results by zero-shot code prompting on coin flip.}
\end{table}

\section{Error Analysis for symbolic reasoning}\label{appendix:error_symbolic}
Here we show an example to illustrate how ambiguity disturbs CoT prompting.
    \begin{table}[H]
        \begin{tabularx}{\textwidth}{X}
        \toprule
        Q: A coin is heads up. Christian doesn't flip the coin. Lewis doesn't flip the coin. Becker flips the coin. Is the coin still heads up? Note that ``flip'' here means ``reverse''. \\
        \\
        A: Let's think step by step.\\
        1. The coin is heads up.\\
        2. Christian doesn't flip the coin, so it remains heads up.\\
        3. Lewis doesn't flip the coin, so it still remains heads up.\\
        4. Becker flips the coin. We don't know the result of the flip, so we can't say for sure whether the coin is still heads up or not.\\
        Therefore, we can't determine whether the coin is still heads up or not without knowing the result of Becker's flip. Therefore, the answer (Yes or No) is unknown. \\\bottomrule     
        \end{tabularx}
        \setlength{\abovecaptionskip}{0.3cm}
        \caption{Example of ambiguity in CoT prompting. In the specific example, in the fourth step, the LLM is confused by the meaning of ``flip'' and thus fail to give the correct answer.}
        \label{table:disambiguation}
    \end{table}

\section{Experiment details for arithmetic reasoning}\label{appendix:exp_arithmetic}
Here we show full prompts for zero-shot CoT prompting, zero-shot code prompting (+irr/+equ), few-shot CoT, PAL, few-shot code prompting (+irr/+equ) and few-shot code prompting + LLM self-contained. Furthermore, we show the results for few-shot code prompting + LLM self-contained.

Besides, we add system messages for all the prompting methods to align with~\citet{pal}. For CoT prompting, we set the system message to ``You will solve math problems.''; for PAL and code prompting, we set the system message to ``You will write python program to solve math problems. You will only write code blocks.''.

\textbf{Zero-shot CoT}
\begin{table}[H]
\begin{tabular}{l}
\toprule
Q: \{question\}              \\
A: Let's think step by step. \\\bottomrule
\end{tabular}
\caption{Zero-shot CoT prompt for math world problems.}
\end{table}

\clearpage
\textbf{Zero-shot code}
\begin{table}[H]
\begin{tabular}{l}
    \toprule
\begin{tabular}[c]{@{}l@{}}{\ul \textbf{plain:}}\\Generate python code to answer the   question.\\Note that code should follow the format   \Verb|`|\Verb|`|\Verb|`|code\Verb|`|\Verb|`|\Verb|`|.\\Q: \{question\}\end{tabular}  \\\midrule
\begin{tabular}[c]{@{}l@{}}{\ul \textbf{+irr:}}\\Generate python code to answer the   question.\\Note that code should follow the format   \Verb|`|\Verb|`|\Verb|`|code\Verb|`|\Verb|`|\Verb|`|.\\There may be irrelevant information in   the question. If you find it, ignore it.\\Q: \{question\}\end{tabular}\\\midrule
\begin{tabular}[c]{@{}l@{}}{\ul \textbf{+equ:}}\\Generate python code to answer the   question.\\Note that code should follow the format   \Verb|`|\Verb|`|\Verb|`|code\Verb|`|\Verb|`|\Verb|`|.\\If you need to solve an equation,   here's an instruction:\\\Verb|`|\Verb|`|\Verb|`|python\\\# to solve an equation, you can use   python package sympy\\import sympy\\     \# for example, to solve 2*x = 5\\\# First, declarify your variable, in this case, 'x'\\x = sympy.symbols("x")\\     \# Second, transform the equation so that the right hand side of the   equation is zero.\\\# 2*x - 5 = 0\\     \# Third, use `sympy.solve` to solve the equation\\a = sympy.solve({[}2 * x - 5{]}, {[}x{]})\\     \# Print the output as a float. Note that `a` is a dict\\print(float(a{[}x{]}))\\\Verb|`|\Verb|`|\Verb|`|\\     Q: \{question\}\end{tabular}\\\bottomrule
\end{tabular}
\caption{Zero-shot code prompt (plain/+irr/+equ) for math world problems.}
\end{table}

\clearpage
\textbf{Few-shot CoT}

\begin{figure}[H]
    \centering
    \includegraphics[width=\textwidth]{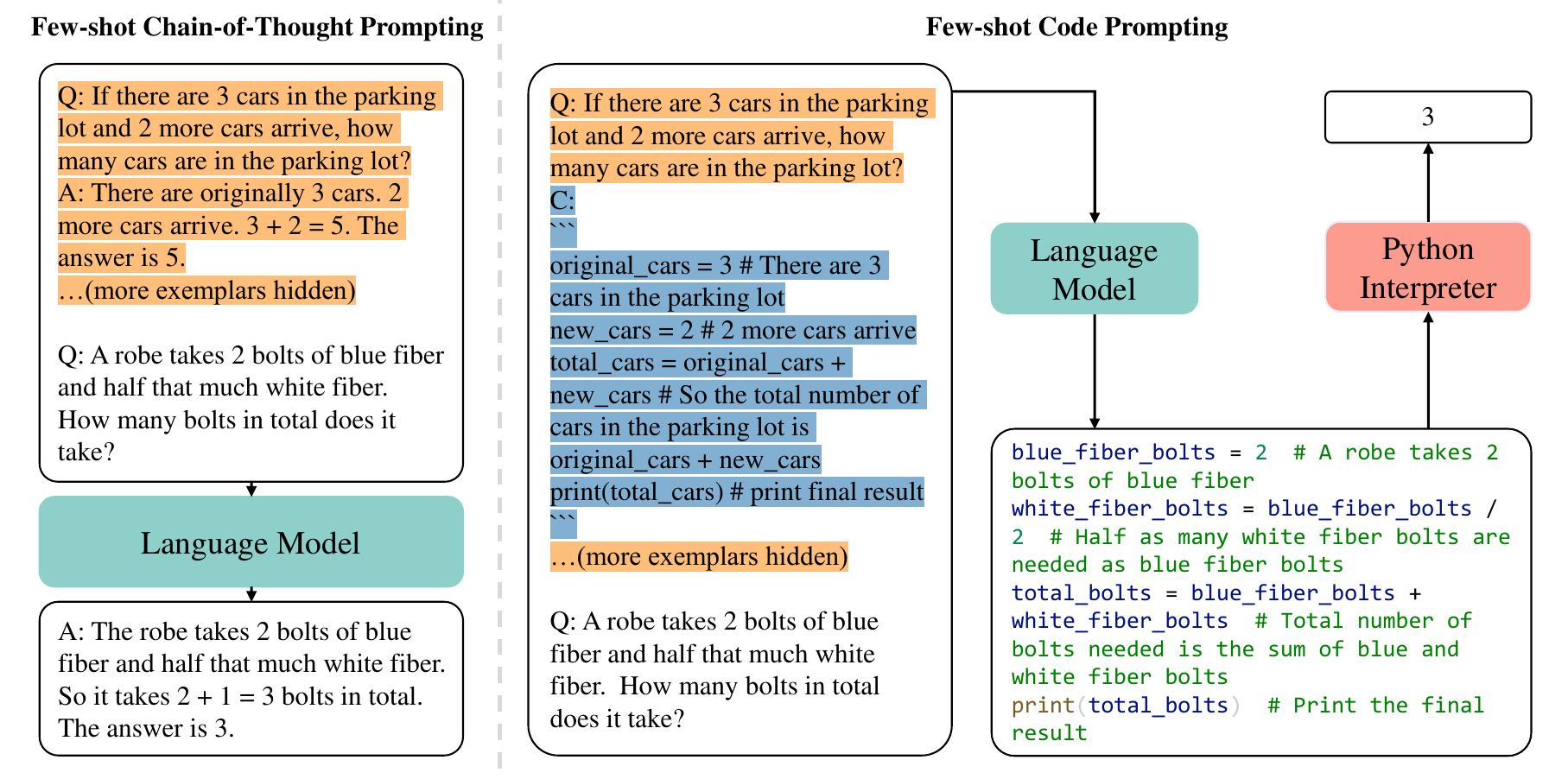}
    \caption{The pipelines of few-shot CoT prompting and few-shot code prompting are shown in the figure.}
    \label{fig:few-shot pipeline}
\end{figure}

\begin{table}[H]
\begin{tabularx}{\textwidth}{X}
    \toprule
Let's think step by step to solve math problems. Here are three examples how to do it,\\ \\
Q: Olivia has \$23. She bought five bagels for \$3 each. How much money does she have left? \\
A: Olivia had 23 dollars. 5 bagels for 3 dollars each will be 5 x 3 = 15 dollars. So she has 23 - 15 dollars left. 23 - 15 is 8. The answer is 8.                     \\ \\
Q: Michael had 58 golf balls. On tuesday, he lost 23 golf balls. On wednesday, he lost 2 more. How many golf balls did he have at the end of wednesday?               \\
A: Michael started with 58 golf balls. After losing 23 on tuesday, he had 58 - 23 = 35. After losing 2 more, he had 35 - 2 = 33 golf balls. The answer is 33.  \\ \\
Q: There were nine computers in the server room. Five more computers were installed each day, from monday to thursday. How many computers are now in the server room? \\
A: There were originally 9 computers. For each of 4 days, 5 more computers were added. So 5 * 4 = 20 computers were added. 9 + 20 is 29. The answer is 29.            \\ \\
How about this question? \\
Q: \{question\} \\\bottomrule           
\end{tabularx}
\caption{Few-shot CoT prompt for math world problems. We follow the code released by~\citet{pal} to choose the exemplars.}
\end{table}

\clearpage
\textbf{PAL}
\begin{table}[H]
\begin{tabularx}{\textwidth}{X}
\toprule
Let's use python to solve math problems. Here are three examples how to do it,    \\ \\
Q: Olivia has \$23. She bought five bagels for \$3 each. How much money does she have left?       \\
\Verb|`|\Verb|`|\Verb|`|    \\
def solution():       \\
"""Olivia has \$23. She bought five bagels for \$3 each. How much money does she have left?"""    \\
money\_initial = 23   \\
bagels = 5            \\
bagel\_cost = 3       \\
money\_spent = bagels * bagel\_cost          \\
money\_left = money\_initial - money\_spent  \\
result = money\_left  \\
return result         \\
\Verb|`|\Verb|`|\Verb|`|    \\ \\
Q: Michael had 58 golf balls. On tuesday, he lost 23 golf balls. On wednesday, he lost 2 more. How many golf balls did he have at the end of wednesday?     \\
\Verb|`|\Verb|`|\Verb|`|    \\
def solution():       \\
"""Michael had 58 golf balls. On tuesday, he lost 23 golf balls. On wednesday, he lost 2 more. How many golf balls did he have at the end of wednesday?"""  \\
golf\_balls\_initial = 58      \\
golf\_balls\_lost\_tuesday = 23              \\
golf\_balls\_lost\_wednesday = 2             \\
golf\_balls\_left = golf\_balls\_initial - golf\_balls\_lost\_tuesday - golf\_balls\_lost\_wednesday     \\
result = golf\_balls\_left     \\
return result         \\
\Verb|`|\Verb|`|\Verb|`|    \\ \\
Q: There were nine computers in the server room. Five more computers were installed each day, from monday to thursday. How many computers are now in the server room?     \\
\Verb|`|\Verb|`|\Verb|`|    \\
def solution():       \\
"""There were nine computers in the server room. Five more computers were installed each day, from monday to thursday. How many computers are now in the server room?"""  \\
computers\_initial = 9         \\
computers\_per\_day = 5        \\
num\_days = 4        \# 4 days between monday and thursday \\
computers\_added = computers\_per\_day * num\_days         \\
computers\_total = computers\_initial + computers\_added   \\
result = computers\_total      \\
return result         \\
\Verb|`|\Verb|`|\Verb|`|    \\ \\
How about this question?       \\
Q: \{question\}      \\\bottomrule
\end{tabularx}
\caption{PAL prompt for math world problems. We use the demonstrations from the code released by~\citet{pal}.}
\end{table}

\clearpage
\textbf{Few-shot code}
\begin{table}[H]
\begin{tabularx}{\textwidth}{X}
    \toprule
Let's use python to solve math   problems. Here are three examples how to do it,\\ \\ Q: Olivia has \$23. She bought five   bagels for \$3 each. How much money does she have left?\\ \Verb|`|\Verb|`|\Verb|`|\\ def solution():\\     """Olivia   has \$23. She bought five bagels for \$3 each. How much money does she have   left?"""\\     money\_initial = 23 \# Olivia has \$23 initially\\     bagels = 5 \# Olivia   bought 5 bagels\\     bagel\_cost = 3 \# Each bagel cost  \$3\\     money\_spent = bagels * bagel\_cost \# The total cost of 5 bagels is the   product of the price of each bagel and the number of bagels\\     money\_left = money\_initial - money\_spent \# Money left is the difference between   initial money and the total cost of 5 bagels\\     result = money\_left\\     return result\\ \Verb|`|\Verb|`|\Verb|`|\\ \\Q: Michael had 58 golf balls. On tuesday, he lost 23 golf balls. On wednesday, he lost 2 more. How many golf balls   did he have at the end of wednesday?\\ \Verb|`|\Verb|`|\Verb|`|\\ def solution():\\     """Michael   had 58 golf balls. On tuesday, he lost 23 golf balls. On wednesday, he lost 2 more. How many golf balls   did he have at the end of wednesday?"""\\     golf\_balls\_initial = 58 \# Michael had 58 golf balls   initially\\     golf\_balls\_lost\_tuesday = 23 \# He lost 23 golf balls on   Tuesday\\     golf\_balls\_lost\_wednesday = 2 \# He lost 2 more golf balls on   Wednesday\\     golf\_balls\_left = golf\_balls\_initial - golf\_balls\_lost\_tuesday - golf\_balls\_lost\_wednesday \# Number of golf balls remaining is   the difference between initial golf balls and the golf balls lost on Tuesday   and Wednesday\\     result = golf\_balls\_left\\     return result\\ \Verb|`|\Verb|`|\Verb|`|\\  \\Q: There were nine computers in the server room. Five more computers were   installed each day, from monday to thursday. How many computers are now in the   server room?\\ \Verb|`|\Verb|`|\Verb|`|\\ def solution():\\     """There   were nine computers in the server room. Five more computers were installed   each day, from monday to thursday. How many computers are now in the   server room?"""\\     computers\_initial = 9 \# There were 9 computers in the   server room initially\\     computers\_per\_day = 5 \# 5 more computers were installed   each day from Monday to Thursday\\     num\_days = 4  \# 4 days between monday and thursday\\     computers\_added = computers\_per\_day * num\_days \# Additional computers are the product   of the number of computers installed each day and the number of days\\     computers\_total = computers\_initial + computers\_added \# Total number of computers is the sum   of initial computers and the additional computers installed on 4 days\\     result = computers\_total\\     return result\\ \Verb|`|\Verb|`|\Verb|`|\\ \\  How about this question?\\Q: \{question\}\\\bottomrule 
\end{tabularx}
\caption{Few-shot code prompt for math world problems. Here we present the version of adding annotations at the end of each line of the code.}
\end{table}

\clearpage
\textbf{Few-shot code + LLM self-contained}
\begin{table}[H]
\begin{tabularx}{\textwidth}{X}
    \toprule
Let's use python to solve math problems. Here are three examples how to do it,\\ \\
Q: Olivia has \$23. She bought five bagels for \$3 each. How much money does she have left?
\\ \Verb|`|\Verb|`|\Verb|`|\\
def solution():\\
    """Olivia has \$23. She bought five bagels for \$3 each. How much money does she have left?"""\\
    money\_initial = 23 \# Olivia has \$23 initially\\
    bagels = 5 \# Olivia bought 5 bagels\\
    bagel\_cost = 3 \# Each bagel cost \$3\\
    money\_spent = bagels * bagel\_cost \# The total cost of 5 bagels is the product of the price of each bagel and the number of bagels\\
    money\_left = money\_initial - money\_spent \# Money left is the difference between initial money and the total cost of 5 bagels\\
    result = money\_left\\
    return result
\\ \Verb|`|\Verb|`|\Verb|`|\\
Let's think step by step according to the code:\\
\\ \Verb|`|\Verb|`|\Verb|`|\\
money\_initial = 23\\
bagels = 5\\
bagel\_cost = 3\\
money\_left = money\_initial - money\_spent = 23 - 15 = 8\\
result = money\_left = 8
\\ \Verb|`|\Verb|`|\Verb|`|\\
So the answer is 8.
\\ \\
Q: Michael had 58 golf balls. On tuesday, he lost 23 golf balls. On wednesday, he lost 2 more. How many golf balls did he have at the end of wednesday?
\\ \Verb|`|\Verb|`|\Verb|`|\\
def solution():\\
    """Michael had 58 golf balls. On tuesday, he lost 23 golf balls. On wednesday, he lost 2 more. How many golf balls did he have at the end of wednesday?"""\\
    golf\_balls\_initial = 58 \# Michael had 58 golf balls initially \\
    golf\_balls\_lost\_tuesday = 23 \# He lost 23 golf balls on Tuesday \\
    golf\_balls\_lost\_wednesday = 2 \# He lost 2 more golf balls on Wednesday\\
    golf\_balls\_left = golf\_balls\_initial - golf\_balls\_lost\_tuesday - golf\_balls\_lost\_wednesday \# Number of golf balls remaining is the difference between initial golf balls and the golf balls lost on Tuesday and Wednesday\\
    result = golf\_balls\_left\\
    return result
\\ \Verb|`|\Verb|`|\Verb|`|\\
Let's think step by step according to the code:
\\ \Verb|`|\Verb|`|\Verb|`|\\
golf\_balls\_initial = 58\\
golf\_balls\_lost\_tuesday = 23\\
golf\_balls\_lost\_wednesday = 2\\
golf\_balls\_left = golf\_balls\_initial - golf\_balls\_lost\_tuesday - golf\_balls\_lost\_wednesday = 58 - 23 - 2 = 33\\
result = golf\_balls\_left = 33\\
\\ \Verb|`|\Verb|`|\Verb|`|\\
So the answer is 33.\\\bottomrule
\end{tabularx}
\caption{Prompt of few-shot code + LLM self-contained for math world problems.}
\end{table}

\begin{table}[H]
\begin{tabularx}{\textwidth}{X}
    \toprule
Q: There were nine computers in the server room. Five more computers were installed each day, from monday to thursday. How many computers are now in the server room?
\\ \Verb|`|\Verb|`|\Verb|`|\\
def solution():\\
    """There were nine computers in the server room. Five more computers were installed each day, from monday to thursday. How many computers are now in the server room?"""\\
    computers\_initial = 9 \# There were 9 \\computers in the server room initially
    computers\_per\_day = 5 \# 5 more computers were installed each day from Monday to Thursday\\
    num\_days = 4  \# 4 days between monday and thursday\\
    computers\_added = computers\_per\_day * num\_days \# Additional computers are the product of the number of computers installed each day and the number of days\\
    computers\_total = computers\_initial + computers\_added \# Total number of computers is the sum of initial computers and the additional computers installed on 4 days\\
    result = computers\_total
    return result
\\ \Verb|`|\Verb|`|\Verb|`|\\
Let's think step by step according to the code:\\
\\ \Verb|`|\Verb|`|\Verb|`|\\
computers\_initial = 9\\
computers\_per\_day = 5\\
num\_days = 4\\
computers\_added = computers\_per\_day * num\_days = 5 * 4 = 20\\
computers\_total = computers\_initial + computers\_added = 9 + 20 = 29\\
result = computers\_total = 29
\\ \Verb|`|\Verb|`|\Verb|`|\\
So the answer is 29.\\
\\
How about this question?\\Q: \{question\}\\\bottomrule 
\end{tabularx}
\end{table}

The results of self-contained LLM reasoning after generating the code for arithmetic tasks are shown in Table~\ref{table: few-shot_code_self-contained}. While \citet{pal} claims that ``while current LLMs can be excellent at specifying a high-level plan to solve a task—they are still incapable of executing them'', we show through experiments that external tools like Python interpreter may not be necessary, and that symbolic and arithmetic reasoning abilities are self-contained for LLMs. 

\begin{table}[H]
    \centering
    \begin{tabular}{cccccc}
    \toprule[1pt]
        & SingleEq & AddSub & SVAMP & MultiArith  & GSM8K\\\midrule
    few-shot code + LLM self-contained  & 97.44 & 90.89 & 79.80 & 96.83 & 77.62\\
    \bottomrule[1pt]
    \end{tabular}
    \setlength{\abovecaptionskip}{0.3cm}
    \caption{The accuracy (\%) of few-shot code prompting (+ LLM self-contained) on arithmetic reasoning datasets.
    }
    \label{table: few-shot_code_self-contained}
\end{table}

\section{Error analysis for arithmetic reasoning}
\label{appendix:error_arithmetic}
\subsection{AddSub}
As mentioned in~\S\ref{error analysis}, in AddSub, LLMs make errors mainly due to the distraction of irrelevant variables. Here we show two examples of the LLMs distracted by irrelevant variables.

\begin{table}[H]
  \centering
  \begin{tabularx}{\textwidth}{X|X}
    \toprule
    Q: Joan had 695 Pokemon cards, and 6 were torn. Sara bought 133 of Joan's Pokemon cards. How many Pokemon cards does Joan have now?  
    & 
    Q:  Dan spent \$11.76 on a snake toy, and a cage cost him \$14.54. Dan also found a dollar bill on the ground. What was the total cost of Dan's purchases? 
    \\
    &
    \\
    \begin{minipage}[b]{0.49\columnwidth}
		\centering
		\raisebox{.1\height}{\includegraphics[width=\linewidth]{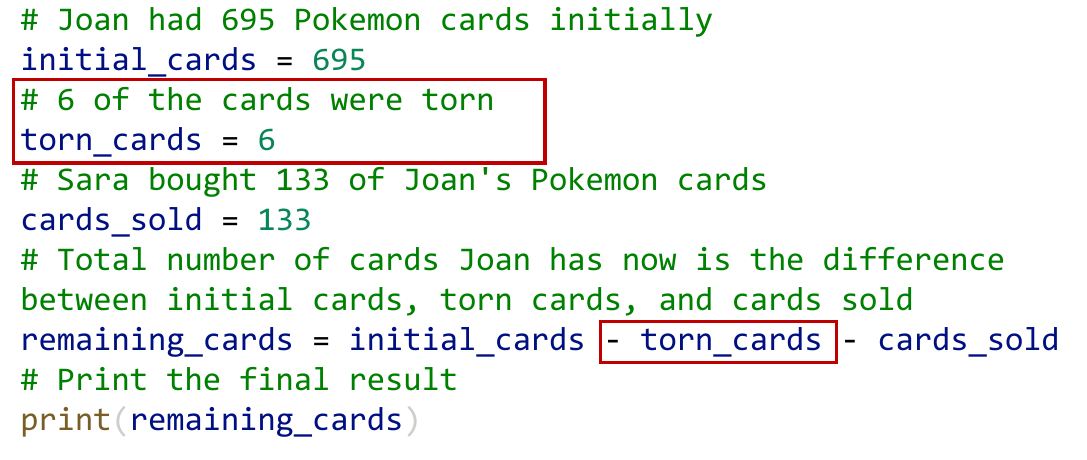}}
    \end{minipage}
    &
    \begin{minipage}[b]{0.49\columnwidth}
		\centering
		\raisebox{.5\height}{\includegraphics[width=\linewidth]{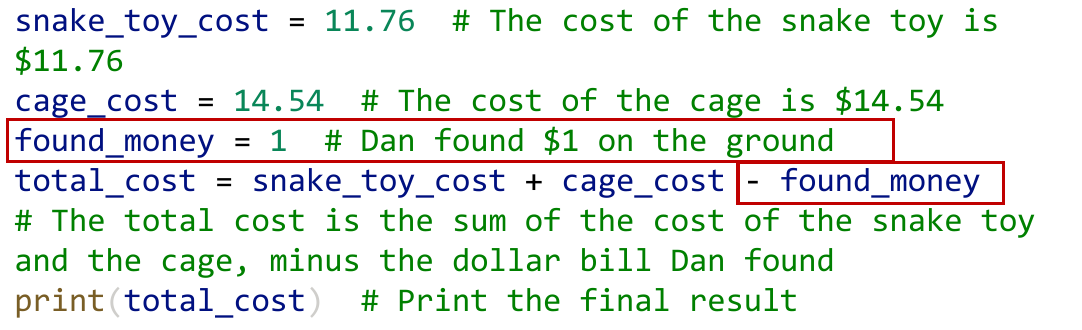}}
    \end{minipage}
    \\ \bottomrule
  \end{tabularx}
  \setlength{\abovecaptionskip}{0.3cm}
  \caption{Error examples for AddSub. In the left example, \texttt{torn\_card} is an irrelevant variable that should not appear in the equation. The correct answer should be \texttt{remaining\_cards = initial\_cards - torn\_cards}, which gives \(562\). In the right example, \texttt{found\_money} is an irrelevant variable. The correct answer should be \texttt{total\_cost = snake\_toy\_cost + cage\_cost}, which gives \(26.3\).}
  \label{table:addsub}
\end{table}

\subsection{MultiArith}
We have mentioned in~\S\ref{error analysis} that in MultiArith, the wrong answers from code prompting are mainly due to two reasons: 1) Code prompting is more sensitive to ambiguity in the question, which may leads to a wrong answer.  2) The groundtruth for some questions is wrong, while the predictions are correct.

Here we show examples for both cases.

\begin{table}[H]
  \centering
  \begin{tabularx}{\textwidth}{X}
    \toprule
    Q:  The school cafeteria ordered 43 red apples and 32 green apples for students lunches. But, if only 2 students wanted fruit, how many extra did the cafeteria end up with?  
    \\
    \\
    \begin{minipage}[b]{\columnwidth}
		\centering
		{\includegraphics[width=\linewidth]{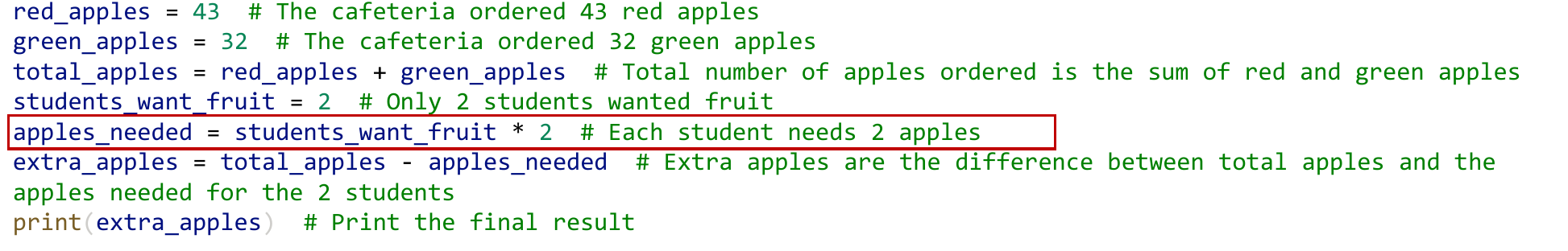}}
    \end{minipage}
    \\ \bottomrule
  \end{tabularx}
  \setlength{\abovecaptionskip}{0.3cm}
  \caption{An error example of few-shot code prompting for MultiArith. The question does not make it clear how many apple a student want, but the ground truth suggests that we should take it as default that each student wants \(1\) apple. Code prompting makes the LLM more sensitive to such ambiguity.}
  \label{table:multiarith_1}
\end{table}

\begin{table}[H]
  \centering
  \begin{tabularx}{\textwidth}{X}
    \toprule
    Q:  Emily's team won their dodgeball game and scored 39 points total. If Emily scored 23 of the points and everyone else scored 2 points each, how many players were on her team? 
    \\
    \\
    \begin{minipage}[b]{\columnwidth}
		\centering
		{\includegraphics[width=\linewidth]{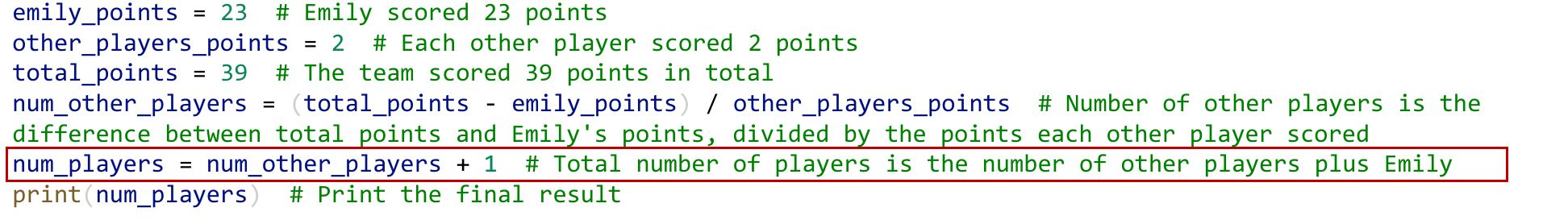}}
    \end{minipage}
    \\ \bottomrule
  \end{tabularx}
  \setlength{\abovecaptionskip}{0.3cm}
  \caption{An error example of few-shot code prompting for MultiArith. The ground truth fails to count Emily into the players, resulting in wrong answer \((39-23)/2=8\).}
  \label{table:multiarith_2}
\end{table}

\subsection{GSM8K}

\textbf{Error with equations.}
We have mentioned in~\S\ref{error analysis} that the LLM struggles to solve equations in code. 

When facing an equation, the LLM tends to solve it in the annotations or directly lists the equation in the code, which may easily result in wrong answers or bugs. Here are two examples.
    
\begin{table}[H]
  \centering
  \begin{tabularx}{\textwidth}{X}
    \toprule
    Q:  Ruby is 6 times older than Sam. In 9 years, Ruby will be 3 times as old as Sam. How old is Sam now? 
    \\
    \\
    \begin{minipage}[b]{\columnwidth}
		{\includegraphics[width=0.6\linewidth]{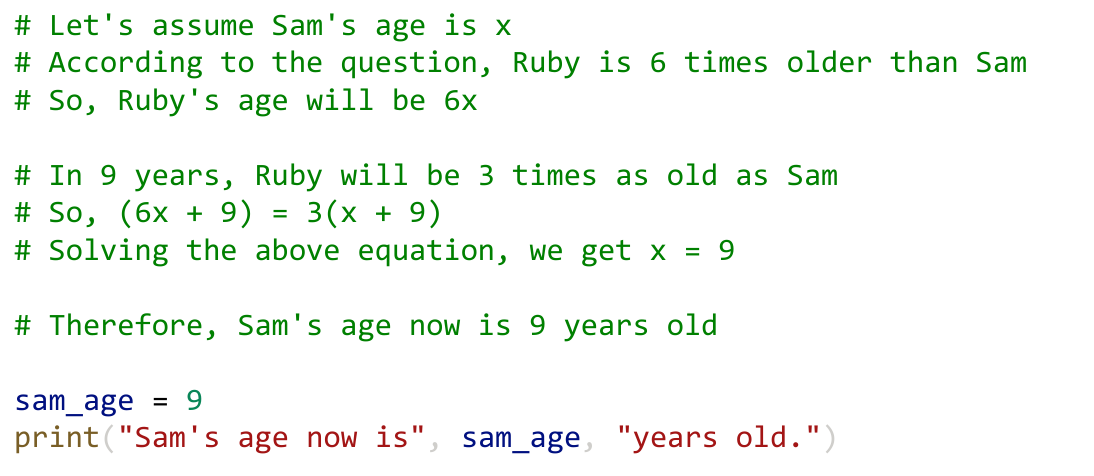}}
    \end{minipage}
    \\ \bottomrule
  \end{tabularx}
  \setlength{\abovecaptionskip}{0.3cm}
  \caption{An error example of code prompting for GSM8K.}
  \label{table:gsm8k_1}
\end{table}

\begin{table}[H]
  \centering
  \begin{tabularx}{\textwidth}{X}
    \toprule
    Q:  The three angles in a triangle add up to 180 degrees. One angle is twice the smallest angle, and one angle is three times the smallest angle. What is the measure of the largest angle in the triangle, in degrees?
    \\
    \\
    \begin{minipage}[b]{\columnwidth}
		{\includegraphics[width=0.6\linewidth]{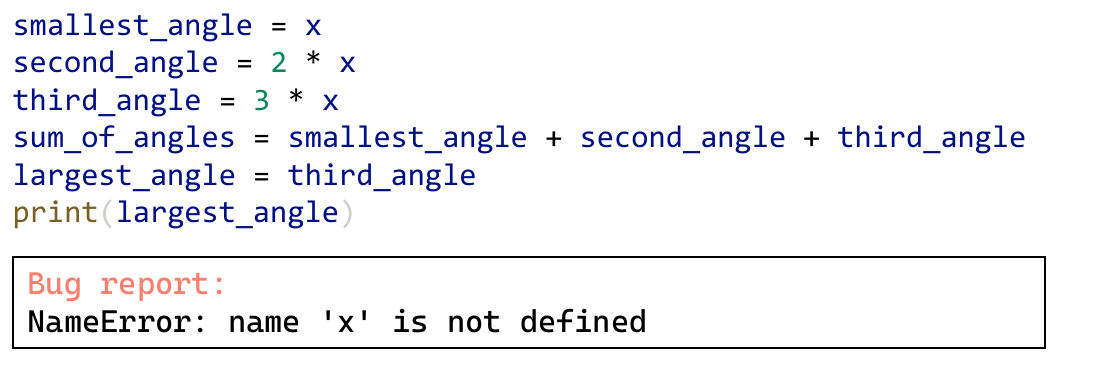}}
    \end{minipage}
    \\ \bottomrule
  \end{tabularx}
  \setlength{\abovecaptionskip}{0.3cm}
  \caption{An error example of code prompting for GSM8K.}
  \label{table:gsm8k_2}
\end{table}

\textbf{Error distribution of CoT prompting and code prompting.}
Here we show the error distribution of few-shot CoT prompting and few-shot code prompting(w/ annotations at the end). Since here we only investigate into the error overlap of CoT prompting and code prompting, to eliminate the effects of system messages and instructions, we remove all the system messages and unnecessary instructions (They are added in the former experiments to align with the method PAL~\citet{pal}). Besides, we use the original 8 exemplars from~\citet{few-shot_cot}. In~\S\ref{ensemble}, we show the results of 8-shot CoT prompting, 8-shot code prompting and the ensemble of them.

\begin{figure}[H]
    \centering
    \includegraphics[width=.3\textwidth]{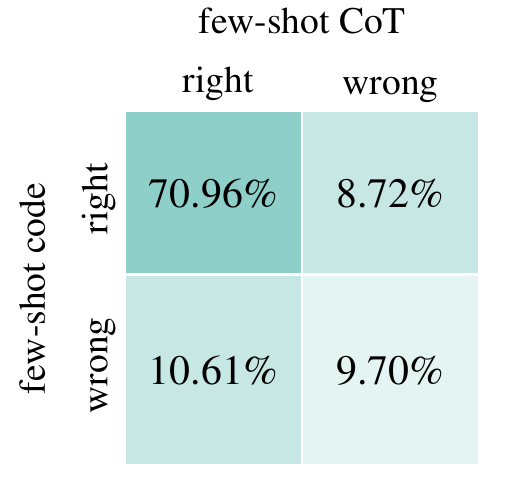}
    \caption{Error distribution of few-shot code prompting and few-shot CoT prompting regarding dataset GSM8K.}
    \label{fig:error distribution}
\end{figure}
\end{document}